\documentclass{article}

\usepackage{microtype}
\usepackage{graphicx}
\usepackage{subcaption}
\usepackage{booktabs}
\usepackage{multirow}
\usepackage{makecell}
\usepackage{hyperref}

\providecommand{\blindsubmission}{0}
\ifnum\blindsubmission=1
  \usepackage{icml2026}
\else
  \usepackage[preprint]{icml2026}
\fi
\usepackage{iftex}
\ifXeTeX\usepackage[T1]{fontenc}\fi

\usepackage{amsmath}
\usepackage{amssymb}
\usepackage{mathtools}
\usepackage{amsthm}
\usepackage[capitalize,noabbrev]{cleveref}

\usepackage{tikz}
\usetikzlibrary{positioning,arrows.meta,fit,calc,decorations.pathreplacing}
\usepackage{pgfplots}
\usepackage[breakable]{tcolorbox}  
\pgfplotsset{compat=1.17}

\usepackage[textsize=tiny]{todonotes}

\usepackage{xspace}
\usepackage{xcolor}

\newcommand{\eyejev}{RetiJev\xspace}
\newcommand{\data}{RetiJev-Data\xspace}          
\newcommand{\bench}{RetiJev-Bench\xspace}        
\newcommand{\jev}{Jev~1.13\xspace}              
\newcommand{\medjev}{MedJev\xspace}
\newcommand{\qwenbig}{Qwen3.8-27B\xspace}

\ifnum\blindsubmission=1
  \newcommand{\releaseat}{upon publication}
\else
  \newcommand{\releaseat}{at \url{https://github.com/Awenbocc/RetiJev}}
\fi

\newcommand{\noul}{\texttt{noul}\xspace}
\newcommand{\choice}{\texttt{choice}\xspace}
\newcommand{\score}{\texttt{score}\xspace}

\newcommand{\exbox}[1]{\begin{tcolorbox}[breakable,colback=black!2,colframe=black!40,boxrule=0.4pt,arc=1pt,
  left=4pt,right=4pt,top=3pt,bottom=3pt,fontupper=\small]#1\end{tcolorbox}}

\newcommand{\nPMCFullTexts}{3{,}868{,}180\xspace}
\newcommand{\nDevQ}{9{,}009\xspace}                   
\newcommand{\nCategories}{12\xspace}
\newcommand{\nActions}{60\xspace}                     
\newcommand{\nChoiceActions}{53\xspace}               
\newcommand{\nExtractedRuleRecords}{6{,}303\xspace}    
\newcommand{\nScales}{11\xspace}                      

\newcommand{\nClinicalQ}{78{,}158\xspace}             

\newcommand{\nTestQ}{24{,}662\xspace}                 
\newcommand{\nMainTestQ}{16{,}332\xspace}             
\newcommand{\nReferenceTestQ}{8{,}330\xspace}          
\newcommand{\nGuideQ}{50{,}074\xspace}                

\newcommand{\nTrainRec}{42{,}550\xspace}              
\newcommand{\nTrainQ}{94{,}561\xspace}                
\newcommand{\nExtDiseases}{531\xspace}                
\newcommand{\nExtCases}{6{,}933\xspace}               

\newcommand{\nStateTokMed}{120\xspace}                
\newcommand{\nBranchTokMed}{57\xspace}
\newcommand{\nQPerCase}{2.57\xspace}

\newcommand{\latOursHtwo}{133\,ms\xspace}             
\newcommand{\latOursCPU}{347\,ms\xspace}              
\newcommand{\memOurs}{3.3\,GB\xspace}                 
\newcommand{\latLLMScore}{247\,ms\xspace}             
\newcommand{\latLLMGen}{546\,ms\xspace}              
\newcommand{\latLLMThink}{74.5\,s\xspace}            
\newcommand{\tokThink}{794\xspace}                    

\newcommand{\latNineB}{158\,ms\xspace}

\newcommand{\latNineBKtwenty}{198\,ms\xspace}
\newcommand{\speedupGenNineB}{3.5}                   
\newcommand{\speedupThinkNineB}{470}                 
\newcommand{\latOursKone}{138\,ms\xspace}
\newcommand{\latOursKtwenty}{146\,ms\xspace}
\newcommand{\latSmallKtwenty}{656\,ms\xspace}
\newcommand{\latLLMKtwenty}{1{,}747\,ms\xspace}

\newcommand{\accGradeStd}{99.9\%\xspace}              
\newcommand{\accGradeOOD}{95.9\%\xspace} 
\newcommand{\accGradeOODNineB}{97.3\%\xspace}
\newcommand{\accGradeHard}{89.8\%\xspace} 
\newcommand{\accGradeHardNineB}{96.3\%\xspace}
\newcommand{\gainGradeLow}{10.4\xspace} 
\newcommand{\gainGradeHigh}{23.0\xspace} 
\newcommand{\gainDecNineLow}{5.4\xspace} 
\newcommand{\gainDecNineHigh}{6.5\xspace} 
\newcommand{\accDecUNineB}{84.9\%\xspace} 
\newcommand{\accDecSNineB}{83.9\%\xspace} 
\newcommand{\accDXu}{68.9\%\xspace} 
\newcommand{\accDXuNineB}{84.8\%\xspace} 
\newcommand{\accDXuLLM}{84.8\%\xspace}

\icmltitlerunning{RetiJev: Fast and Source-Grounded Ophthalmic Decision Models}

\ifnum\blindsubmission=0
  \newcounter{@affiluw}
\fi

\begin{document}

\twocolumn[
  \icmltitle{RetiJev: Fast and Source-Grounded Ophthalmic Decision Models}

  \icmlsetsymbol{equal}{*}

  \begin{icmlauthorlist}
    \icmlauthor{Bo Liu}{equal,cipeh,nusoph,uw}
    \icmlauthor{Lianyu Wang}{equal,cipeh,nusoph}
    \icmlauthor{Meng Wang}{cipeh,nusoph}
    \icmlauthor{Ching-Yu Cheng}{cipeh,nusoph,seri,eyeacp}
    \icmlauthor{Huazhu Fu}{astar}
  \end{icmlauthorlist}

  \ifnum\blindsubmission=0
    \setcounter{@affiliationcounter}{6}
  \fi

  \icmlaffiliation{cipeh}{Centre for Innovation and Precision Eye Health, Yong Loo Lin School of Medicine, National University of Singapore, Singapore 119228, Singapore.}
  \icmlaffiliation{nusoph}{Department of Ophthalmology, Yong Loo Lin School of Medicine, National University of Singapore, Singapore 119228, Singapore.}
  \icmlaffiliation{astar}{Institute of Advanced Intelligence and Computing, Agency for Science, Technology
and Research (A*STAR), Singapore 138632, Singapore.}
  \icmlaffiliation{seri}{Singapore Eye Research Institute, Singapore National Eye Centre, Singapore 169856, Singapore.}
  \icmlaffiliation{eyeacp}{Ophthalmology \& Visual Sciences Academic Clinical Program (EYE ACP), Duke-NUS Medical School, Singapore 169856, Singapore.}
  \icmlaffiliation{uw}{University of Washington, Seattle, WA, USA}

  \icmlcorrespondingauthor{Meng Wang}{wang.m@nus.edu.sg}
  \icmlcorrespondingauthor{Ching-Yu Cheng}{chingyu.cheng@nus.edu.sg}
    \icmlcorrespondingauthor{Huazhu Fu}{hzfu@ieee.org}

  \icmlkeywords{clinical decision support, ophthalmology, efficient inference,
    decision models, calibration, synthetic data, benchmark}

  \vskip 0.3in
]

\printAffiliationsAndNotice{\icmlEqualContribution}

\begin{abstract}
Ophthalmic clinical decision support requires fast, accurate, and privacy-preserving inference, yet autoregressive large language models (LLMs) incur substantial latency and computational overhead for decisions among predefined actions. We therefore investigate a decision-oriented alternative to generative reasoning. Specifically, we first introduce \textbf{\data}, a source-grounded ophthalmology corpus comprising \nClinicalQ clinical-decision and diagnosis questions derived from articles in PubMed Central and a MeSH vocabulary covering 531 eye diseases, and \nGuideQ guideline-grading questions spanning \nScales grading systems, along with a \nTestQ-question evaluation benchmark. Based on this, we develop \textbf{\eyejev}, a family of 0.8B--9B decision models that directly score candidate answers without autoregressive decoding.
Across all model sizes, \eyejev outperforms both a 27B LLM and the commercial Jev 1.13 decision model on guideline grading by \gainGradeLow--\gainGradeHigh points. \eyejev-9B further achieves the highest accuracy on clinical decisions, improving over both baselines by \gainDecNineLow--\gainDecNineHigh points while matching them on diagnosis. 
It answers a case in \latNineB on one GPU, \mbox{\speedupGenNineB$\times$} faster than LLM generation and \mbox{\speedupThinkNineB$\times$} faster than reasoning.
Models are 
released \releaseat.
\end{abstract}

\section{Introduction}
\label{sec:intro}

\begin{figure}[t]
  \centering
  \input{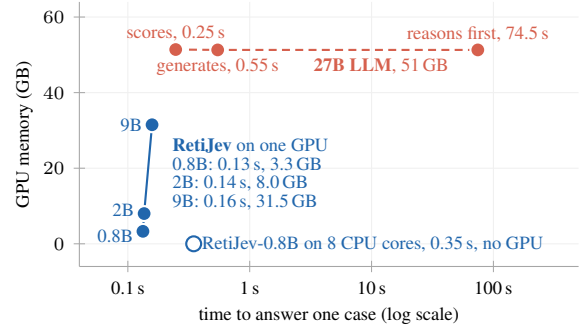}
 \caption{\textbf{Inference latency and peak GPU memory for ophthalmic case-level decision making.}
Latency is measured as the median per-case inference time on a single NVIDIA H200 GPU with batch size 1 (averaging 2.65 questions per case).
All three \eyejev variants achieve 0.13--0.16\,s latency with 3.3--31.5\,GB of peak GPU memory, while \eyejev-0.8B also supports CPU-only inference on eight cores.
The 27B LLM requires approximately 51\,GB of GPU memory and exceeds one minute per case with reasoning enabled.}
  \label{fig:teaser}
\end{figure}

Timely clinical decision-making is critical in ophthalmic care. A patient with sudden painless
vision loss, a red and painful eye after cataract surgery, or a child with a new
squint needs, within minutes, answers to a handful of concrete questions: how
urgently must this be seen, which investigation comes next, is the planned
treatment safe for this patient, and who should manage it? In ophthalmology
many of these calls are made by non-specialists (emergency physicians,
primary-care doctors, optometrists), because eye complaints are frequent,
with roughly two million eye-related emergency visits a year in the United
States \citep{channa2016epidemiology}, while ophthalmologists are
scarce in much of the world \citep{resnikoff2020estimated,burton2021lancet}.
Several eye emergencies are time-critical, such as central retinal artery
occlusion, whose treatment window is measured in hours \citep{macgrory2021crao}. Decision
support helps only if it keeps pace with the encounter. The first of the
``ten commandments'' of clinical decision support is ``speed is everything'' \citep{bates2003ten}.

Large language models (LLMs) have demonstrated substantial
medical and ophthalmic knowledge
\citep{singhal2023large,nori2023capabilities,antaki2023evaluating,thirunavukarasu2024large},
yet their generative paradigm is not well suited to
the practical demands of point-of-care decision support.
Autoregressive decoding introduces latency, particularly
when reasoning or answering multiple questions about
the same patient \citep{wei2022chain,deepseek2025r1}.
LLMs also require substantial computational
resources, while cloud-based inference raises patient
privacy and data governance concerns
\citep{marks2023chatbots,rocher2019estimating}.
Moreover, free-form generation is poorly aligned with
clinical decisions involving predefined actions, and
guideline adherence remains inconsistent
\citep{hager2024evaluation}.

These limitations motivate a different formulation of
point-of-care decision support. Many clinical decisions,
including urgency assessment, investigation selection,
management planning, and guideline grading, involve
\emph{choosing among predefined options rather than generating
free-form responses}. Such tasks can instead be addressed
by directly scoring candidate answers. Decision models
of the Jev family
\citep{palmer2026kev,ma2026medjev,jev2026}
follow this paradigm, producing candidate probabilities
without autoregressive decoding. However, the
general-purpose \jev is available only through a hosted
API, while the open-source \medjev\footnote{\url{https://github.com/JunMa11/MedJev}} focuses on clinical
variable extraction rather than decision-making.
Extending this paradigm to ophthalmology requires
domain-specific supervision. Existing ophthalmic
resources primarily emphasize medical knowledge
assessment or image interpretation, offering limited
source-grounded supervision for clinical decisions
over standardized answer options.

To address this gap, we first introduce \data, a source-grounded ophthalmology corpus comprising two complementary resources. 
The clinical-decision resource contains \nClinicalQ questions derived from articles of PubMed Central (PMC) and a MeSH vocabulary of 531 eye diseases, with labels tied to explicit source evidence.
The guideline-grading resource contains \nGuideQ questions across \nScales published grading systems, with labels derived from clinical criteria and notes independently verified for fidelity to underlying findings. Together, these resources support the systematic evaluation of clinical action selection, guideline application, and disease recognition. 
The test split forms \bench, comprising \nTestQ questions for standardized evaluation. Dataset statistics and task examples are provided in \cref{sec:data,app:example}.

Building on this corpus, we develop \eyejev, a family of
0.8B--9B decision models that directly score candidate
answers without autoregressive decoding. We evaluate
\eyejev against a 27B LLM and the commercial \jev.
Across all model sizes, \eyejev outperforms both baselines
on guideline grading by \gainGradeLow--\gainGradeHigh
percentage points. \eyejev-9B achieves the highest
clinical-decision accuracy while matching the baselines
on diagnosis. As shown in Figure~\ref{fig:teaser}, it answers a case in \latNineB on a single
GPU, compared with \latLLMGen for direct generation and
\latLLMThink for reasoning-based generation by the 27B
LLM. With 34$\times$ fewer parameters,
\eyejev-0.8B outperforms both baselines on seen-decisions and achieves comparable accuracy on unseen-decisions. It requires \memOurs of GPU memory and
runs on eight CPU cores, supporting local
deployment without transmitting patient data to external
services. These results highlight the potential of
compact decision models for accurate, efficient, and
privacy-preserving ophthalmic decision support.

In summary, our contributions are threefold:

\begin{itemize}
  \item We construct \data, a dataset for point-of-care ophthalmic decisions that covers tasks across \nCategories ophthalmic categories and
\nScales guideline-grading systems. 
  \item We propose \eyejev, a family of Jev-style decision models from 0.8B to
  9B parameters that directly score candidate answers without autoregressive decoding.
   \item Extensive experiments show that \eyejev
outperforms a 27B LLM and \jev on guideline
grading at all sizes and on clinical decisions at 2B
and 9B. It achieves 133--158\,ms inference per case
on one GPU, with the 0.8B model also supporting CPU-only
deployment.
\end{itemize}

\section{Related Work}
\label{sec:related}

\paragraph{LLMs for clinical knowledge and decisions.}
LLMs pass licensing-style examinations
\citep{singhal2023large,singhal2025toward,nori2023capabilities} built from
question banks such as MedQA, MedMCQA and PubMedQA
\citep{jin2021disease,pal2022medmcqa,jin2019pubmedqa}. 
Their ophthalmic capabilities have been evaluated
through examination questions and patient queries
\citep{antaki2023evaluating,mihalache2023performance,
thirunavukarasu2024large,bernstein2023comparison},
with further efforts toward domain adaptation
\citep{kim2024leme,gao2023ophglm}.
However, exam accuracy
does not necessarily translate into good decisions: on real inpatient cases LLMs did not
follow guidelines and were sensitive to the order of information
\citep{hager2024evaluation}.
Meanwhile, ophthalmic AI has been mostly image-based, from single-disease screening and
referral systems \citep{gulshan2016development,abramoff2018pivotal,defauw2018clinically}
to retinal foundation models and multimodal benchmarks
\citep{zhou2023retfound,qin2025lmod, wang2026towards}.
Our \eyejev is complementary, addressing text-based clinical
decision support across diverse ophthalmic tasks.

\paragraph{Decision models and efficient inference.}
Existing approaches to efficient inference either
accelerate autoregressive decoding
\citep{kwon2023efficient,leviathan2023fast,zhong2024distserve}
or avoid generation for tasks with predefined answers.
Letter scoring enables LLMs to evaluate candidate
answers directly \citep{robinson2023leveraging},
while encoder-based classifiers provide efficient
discriminative predictions
\citep{devlin2019bert,alsentzer2019publicly,warner2024modernbert}.
More recently, Jev-style decision models directly
score candidate answers without autoregressive decoding
\citep{li2026jev, jev2026,palmer2026kev,li2026specialized}.
Subsequent extensions include \medjev for clinical
variable extraction \citep{ma2026medjev} and
biomedical sentence-encoder approaches to typed
decisions \citep{deka2026typed}.
Building on this paradigm, \eyejev supports
multiple ophthalmic decision tasks by reusing
a shared patient representation and scoring
candidate answers in a single prefill.

\paragraph{Synthetic data and model-verified labels.}
LLM-generated synthetic data has become an established
approach to training language models
\citep{wang2023selfinstruct,gunasekar2023textbooks}.
LLM-based evaluators further enable scalable
verification \citep{zheng2023judging}, although
self-preference bias and correlated errors can
compromise label reliability
\citep{panickssery2024llm}.
To improve supervision quality, we ground clinical
decisions in explicit source evidence and verify
proposed labels across independent model families.
For guideline grading, we instead compute labels
from published clinical rules and independently
verify the fidelity of generated patient descriptions.

\section{\data: Source-Grounded Ophthalmic Decisions}
\label{sec:data}

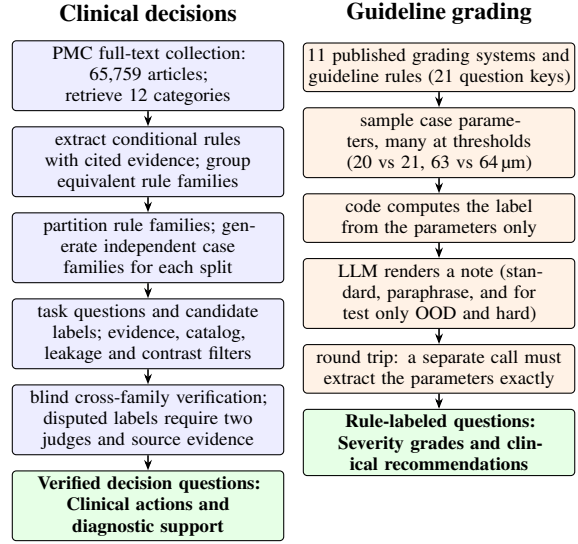
\begin{figure}[t]
  \centering
  \begin{tikzpicture}[
  font=\scriptsize,
  stage/.style={draw, rounded corners=1.5pt, align=center, text width=3.45cm, inner sep=2pt, minimum height=0.62cm},
  dec/.style={stage, fill=blue!7},
  gui/.style={stage, fill=orange!10},
  res/.style={stage, fill=green!10, font=\scriptsize\bfseries},
  arr/.style={-{Stealth[length=1.4mm]}, semithick},
  lab/.style={font=\tiny\itshape, text=black!65, align=center},
  node distance=0.2cm,
]
\node[font=\footnotesize\bfseries] (dt) at (0,0) {Clinical decisions};
\node[dec, below=0.12cm of dt] (d1) {PMC full-text collection: \\ 65{,}759 articles; \\ retrieve \nCategories categories};
\node[dec, below=of d1] (d2) {extract conditional rules with cited evidence; group equivalent rule families};
\node[dec, below=of d2] (d3) {partition rule families; generate independent case families for each split};
\node[dec, below=of d3] (d4) {task questions and candidate labels; evidence, catalog, leakage and contrast filters};
\node[dec, below=of d4] (d5) {blind cross-family verification; disputed labels require two judges and source evidence};
\node[res, below=of d5] (d6) {Verified decision questions: \\ Clinical actions and diagnostic support};
\foreach \a/\b in {d1/d2, d2/d3, d3/d4, d4/d5, d5/d6} { \draw[arr] (\a) -- (\b); }

\node[font=\footnotesize\bfseries] (gt) at (3.85,0) {Guideline grading};
\node[gui, below=0.12cm of gt] (g1) {11 published grading systems and guideline rules (21 question keys)};
\node[gui, below=of g1] (g2) {sample case parameters, many at thresholds (20 vs 21, 63 vs 64\,\textmu m)};
\node[gui, below=of g2] (g3) {code computes the label from the parameters only};
\node[gui, below=of g3] (g4) {LLM renders a note (standard, paraphrase, and for test only OOD and hard)};
\node[gui, below=of g4] (g5) {round trip: a separate call must extract the parameters exactly};
\node[res, below=of g5] (g6) {Rule-labeled questions: \\ Severity grades and clinical recommendations};
\foreach \a/\b in {g1/g2, g2/g3, g3/g4, g4/g5, g5/g6} { \draw[arr] (\a) -- (\b); }
\end{tikzpicture}
  \caption{\textbf{Construction pipeline of \data.}
  Clinical decisions (left): ophthalmic literature is retrieved and organized into evidence bundles.
  Conditional rules guide case synthesis, followed by evidence filtering
  and independent label verification. Guideline grading (right): labels
  are computed from structured case parameters, and independent parameter
  re-extraction verifies the fidelity of generated clinical notes.}
  \label{fig:pipeline}
\end{figure}

We construct \data, a dataset for training and evaluating ophthalmic decision
models. It combines two components: \emph{clinical decisions and diagnosis},
with \nClinicalQ questions from source-grounded and synthetic diagnosis cases;
and \emph{guideline grading}, with \nGuideQ questions for severity assessment
and treatment eligibility based on published criteria. All counts include
the training, development, and test splits. Both components comprise
synthetic patient descriptions paired with questions, candidate answers,
and labels. The following subsections describe the construction and verification of these components (\cref{fig:pipeline}), followed by the evaluation sets that form \bench
(\cref{tab:stats}).

\subsection{Clinical Decision Tasks}
\label{sec:data-decision}

\paragraph{Literature retrieval and evidence organization.}

We retrieved PubMed-indexed literature using queries combining
the names and synonyms of 1{,}817 ophthalmic diseases with
clinical decision terms.
After deduplication and full-text availability screening,
we retained 65{,}759 full-text articles
for evidence extraction and characterized their coverage across
\nCategories{} ophthalmic content categories
(\cref{tab:ophthalmic-categories}).
Guidelines, consensus statements, and clinical reviews
provided explicit clinical recommendations, while original
studies offered complementary evidence on patient populations,
clinical conditions, and outcomes.
Relevant passages were organized into disease-specific
\emph{evidence bundles}, preserving PMCIDs, PMIDs and DOIs
where available, source locations,
and verbatim excerpts for traceability.

\subsubsection{From Articles to Synthetic Cases}
\label{sec:data-questions}

\paragraph{Clinical decision rule extraction.}
An LLM extracts conditional clinical rules from relevant
literature passages, including tables and qualifying
statements. Each rule links patient findings to a supported
diagnosis or clinical action, specifying its applicability
conditions, prerequisites, exceptions, and recommendation
strength. Verbatim source evidence and document identifiers
are retained for traceability. Recommendations are interpreted
according to their original strength, so permissive statements
do not automatically establish preferred actions.
Semantically equivalent rules are grouped across articles
into rule families, while differences that affect clinical
decisions are preserved. Unresolved conflicts are excluded
from unique-answer supervision.
The extraction inventory comprises 6,303 rule
families.

\paragraph{Rule-family partitioning.}
We divide rule families into training, development, and held-out sets before case generation. Equivalent decision principles
extracted from different articles are assigned to the same
pool to prevent cross-split overlap. The \emph{Seen} evaluation cases
are independently generated from rule families represented
in training, whereas \emph{Unseen} cases are generated from held-out
rule families. Thus, Seen evaluates new patient presentations
of familiar decision principles, while Unseen evaluates
principles absent from task-specific training supervision.
Each case family, including its clinical variants,
paraphrases, and derived questions, remains within a single
split. Cases requiring rules from different pools are
excluded to prevent leakage.

\paragraph{Synthetic case generation.}
Based on these rules, an LLM writes related synthetic patient descriptions that differ in a decision-relevant condition, and proposes candidate answers and labels for each case. For example, the retinal pigment epithelial tear article describes characteristic retinal findings and highlights OCT for confirmation. A generated patient presents these findings but has not yet undergone imaging. The generator assigns OCT as the preferred answer to the next-investigation question. Findings that have not been examined remain unknown, and articles without usable decision content yield no cases. More examples can be found in \cref{app:example}.

\subsubsection{Questions and Verified Answers}
\label{sec:data-answers}

\paragraph{Defining tasks and answer options.}
We organize generated questions into six tasks: \emph{referral},
\emph{next investigation}, \emph{management direction}, \emph{plan safety}, \emph{urgency}, and \emph{diagnosis
rating}. The first four \textbf{action} tasks draw from a shared catalog of \nChoiceActions 
options, designed to capture clinical decisions that recur across
diseases at the level of detail supported by source literature. An ophthalmologist
reviewed and approved the scope and granularity of this catalog before case
generation (\cref{tab:catalog}). Defining the catalog in advance constrains
generated options to a clinically reviewed set while allowing cases from
different articles to provide training examples for the same actions. For
each case in these four tasks, the generator selects candidate actions from
the corresponding task's options. The remaining \emph{urgency} and \emph{diagnosis-rating} tasks use four-level and three-level
rating scales, respectively. Including these scales, the complete
answer-option catalog contains \nActions entries.

\paragraph{Constructing questions and labels.}
For the four action tasks, a case may support multiple reasonable actions, even when one is preferred. A single-best-action question captures the preferred choice but loses supervision about acceptable alternatives. For example, in the synthetic case above, OCT is preferred while fundus photography is also acceptable. Some literature may further support several actions without establishing a unique preference. We therefore construct two question types for these tasks: selecting the preferred action and judging whether an individual candidate is acceptable. The latter retains supervision over reasonable alternatives and allows cases without a unique preferred action to contribute training examples.

We derive both question types from a shared assessment of the candidate actions. Given the patient description and source article, the generator first labels each candidate as \emph{preferred} (the best-supported action), \emph{acceptable} (reasonable but not preferred), \emph{inappropriate} (ruled out by the source
or a missing prerequisite), or \emph{unresolved} (not sufficiently supported or excluded by the source).
A single-best-action question is created only
when exactly one candidate is preferred. Almost every such question also
offers clinician deferral if the listed options are insufficient. 
Separately, each preferred or acceptable action yields a binary question labelled \emph{yes}, whereas each inappropriate candidate yields one labelled \emph{no}. 
Unresolved candidates are omitted from binary-question generation, preventing absent source evidence from being treated as negative supervision.

We use fixed templates for each task and question type, inserting the relevant action or diagnosis where needed. This keeps the requested judgment consistent across cases while allowing patient evidence and candidate options to vary. For the last two tasks, urgency questions ask for the case's urgency level, whereas diagnosis questions ask how strongly each candidate diagnosis is supported. The generator assigns answers according to the corresponding rating scales. All proposed labels then undergo the evidence checks and independent verification described below.

\paragraph{Evidence filtering and independent verification.}
We first apply filters to remove unsupported quotes, invalid candidates, answer leakage, and diagnosis ratings that lack supporting case evidence (details in \cref{app:filters}). An LLM from a different family then independently answers each question using the same source article without access to the proposed label. Questions for which the two answers agree form the \emph{agree} subset. Disagreements are retained in the \emph{adjudicated} subset only when two additional model families agree on the same answer, at least one provides a verbatim supporting quote from the source, and neither judges the case to be insufficiently specified. All remaining disagreements are discarded. 
During training and evaluation, the model receives only the patient description, question, and candidate options, without access to the source article.

\subsection{Guideline Grading Task}
\label{sec:data-guideline}

The guideline grading task assesses whether a model can apply published
clinical criteria to a patient description. We construct this component
from \nScales published grading systems and guideline rules (e.g., diabetic
retinopathy severity grading and glaucoma visual-field staging). These
sources define criteria for assigning grades or clinical decisions, which
we apply to synthetically generated patient cases.

\paragraph{From grading rules to synthetic cases.}
Following \medjev \citep{ma2026medjev}, we encode each grading system as
Python functions that compute answers from structured clinical findings.
Cases are generated programmatically by sampling predefined categorical
values and numeric ranges, including signs, measurements, and counts. We
emphasize values near decision thresholds, where small changes can alter
the resulting grade or decision.

An LLM then converts each case into a clinical note without revealing the
computed answer. For example, a right eye with microaneurysms and no other
retinal abnormalities is assigned mild non-proliferative diabetic
retinopathy by the grading function, while the generated note describes
only the underlying findings. Each note is paired with a question and its
code-computed answer. To verify note fidelity, a separate LLM call re-extracts
the structured parameters from each note, and we retain only examples that
exactly recover the parameters used for label computation. We additionally
generate paraphrased notes for the same underlying cases while keeping
labels fixed, to train and evaluate robustness to wording variation
(\cref{app:robust}).

\paragraph{Testing robustness to clinical wording.}
We construct two additional test sets to evaluate robustness to wording and
case difficulty. Specifically, grading-OOD contains notes written by an LLM family not
used to generate the training notes. Grading-Hard combines cases near
decision thresholds with notes containing clinical shorthand and clearly
dated information from earlier visits. Both use grading criteria covered
during training, and all notes pass the same parameter-recovery check
described above. Renderer configurations and the grading-system inventory are
provided in \cref{app:scales,app:grading}.
\begin{table}[t]
  \caption{\textbf{Training data and evaluation sets.}
Counts denote the number of questions. Training data composition and evaluation set definitions are detailed in \cref{tab:training-data,tab:tiers}.}
  \label{tab:stats}
  \centering\small
  \setlength{\tabcolsep}{4pt}
  \begin{tabular*}{\linewidth}{@{\extracolsep{\fill}}lr@{\hspace{1.2em}}lr@{}}
    \toprule
    Training & \nTrainQ & Development & \nDevQ \\
    \midrule
    \multicolumn{4}{@{}l}{\emph{Main test sets (\nMainTestQ questions)}} \\
    Decision-Unseen & 4{,}625 & Decision-Seen & 3{,}106 \\
    Diagnosis-Unseen & 164 & Diagnosis-Seen & 224 \\
    Grading-OOD & 3{,}983 & Grading-Hard & 4{,}230 \\
    \midrule
    \multicolumn{4}{@{}l}{\emph{Grading reference sets (\nReferenceTestQ questions)}} \\
    Grading-Standard & 4{,}167 & Grading-Paraphrase & 4{,}163 \\
    \midrule
    \multicolumn{3}{@{}l}{\textbf{All test questions}} & \textbf{\nTestQ} \\
    \bottomrule
  \end{tabular*}
\end{table}

\subsection{Training Data and Test Benchmark}
\label{sec:data-bench}
\begin{figure*}[t]
  \centering
  \input{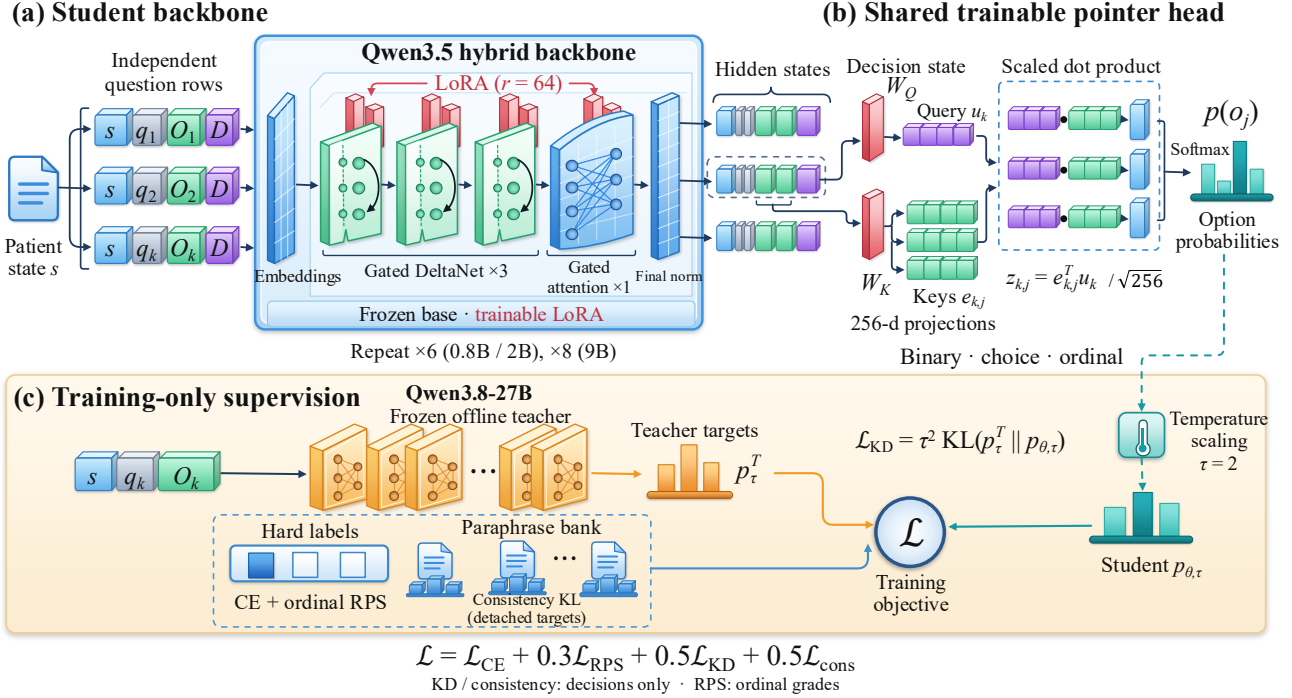}
  \caption{\textbf{\eyejev architecture and training objective.}
  \textbf{(a)} Independent state--question sequences pass through a shared
  backbone with frozen base weights and trainable LoRA adapters on attention,
  DeltaNet and MLP projections. 
  \textbf{(b)} A shared pointer head projects the hidden state at
  $\langle\textsc{d}\rangle$ into a query to score the candidate representations.
  \textbf{(c)} Hard labels supervise all questions, with an additional
  ordinal loss for ordered ratings. Clinical-decision questions can also
  receive offline teacher distributions and paraphrase-consistency targets
  when available. Neither auxiliary target is used for guideline grading.
  Only the adapters and pointer head are trained.}
  \label{fig:arch}
\end{figure*}

\data provides \nTrainQ training and 9{,}009 development
questions, while its \nTestQ-question test split defines
\bench (\cref{tab:stats}). \bench comprises six main test sets
(\nMainTestQ questions) covering decision making (Decision),
diagnosis selection (Diagnosis), and guideline grading (Grading), along with two
grading reference sets (\nReferenceTestQ questions). Note that diagnosis selection follows the single-best-action formulation, converting verified per-candidate diagnosis ratings into a single-best-diagnosis question only when exactly one candidate is labeled as the leading diagnosis. 
Task definitions and test conditions are detailed in \cref{app:bench}.

\section{\eyejev: A Single-Pass Decision Model}
\label{sec:method}

Given a patient description, a question, and a set of
candidate answers, \eyejev learns to score these candidates directly, as shown in \cref{fig:arch}.
Based on the \medjev \citep{ma2026medjev}, we use a shared pointer head across all question types and design a unified training objective that combines label supervision, teacher distributions,
and paraphrase consistency. At inference, multiple questions reuse the same patient representation while remaining independent of one another, thereby improving efficiency.

\subsection{Problem Formulation and Candidate Scoring}
\label{sec:method-model}

\Cref{fig:arch}(a,b) illustrates the shared backbone and candidate-scoring
head. A case consists of a patient state $s$ and $K$ questions. Each question
$q_k$ contains an instruction and a candidate list
$O_k=(o_{k,1},\dots,o_{k,m_k})$. Candidates are actions or diagnoses for
choice questions, (no, yes) for binary judgments, and ordered levels for
rating or grading questions. The model predicts
$p_\theta(o_{k,j}\mid s,q_k)$ over the candidates of each question.
Predictions depend only on the state and the question being answered, so
adding, removing or reordering other questions leaves them unchanged.

\paragraph{Shared scoring head.}
Candidate sets vary across tasks and cases. To accommodate this variation,
we score candidate representations with a pointer head
\citep{vinyals2015pointer}, rather than assign a separate classifier to each
task. 
As shown in \cref{fig:arch}(a), the input is constructed by concatenating the patient
state $s$, the question instruction $q_k$, and the
candidate texts in $O_k$, followed by a decision token
$\langle\textsc{d}\rangle$ (denoted as $D$ in the figure).
The hidden state at this token position serves as the
decision representation.
A causal backbone encodes the sequence, so the
hidden state at this final token incorporates the preceding state,
instruction and all candidates. Each
candidate ends with an option-closing token
$\langle/\textsc{o}\rangle$, whose hidden state represents that candidate
in context. Learned affine projections map the hidden state at
$\langle\textsc{d}\rangle$ to a query $u_k$ and the hidden state at each
$\langle/\textsc{o}\rangle$ to a key $e_{k,j}$, both of dimension $d_h$
(\cref{fig:arch}(b)). The answer distribution is
\begin{equation}
  p_\theta(o_{k,j}\mid s,q_k)
  =\operatorname{softmax}_j\!\left(
    \frac{e_{k,j}^{\top}u_k}{\sqrt{d_h}}\right).
  \label{eq:pointer}
\end{equation}
The same head therefore supports different numbers and meanings of
candidates without changing its parameters. The predicted answer is the
highest-probability candidate. The backbone is Qwen3.5 \citep{qwen2026qwen35}, which
combines attention and Gated DeltaNet layers \citep{yang2025gated}.
Input details are given in \cref{app:encoding}.

\begin{figure*}[t]
  \centering
  \input{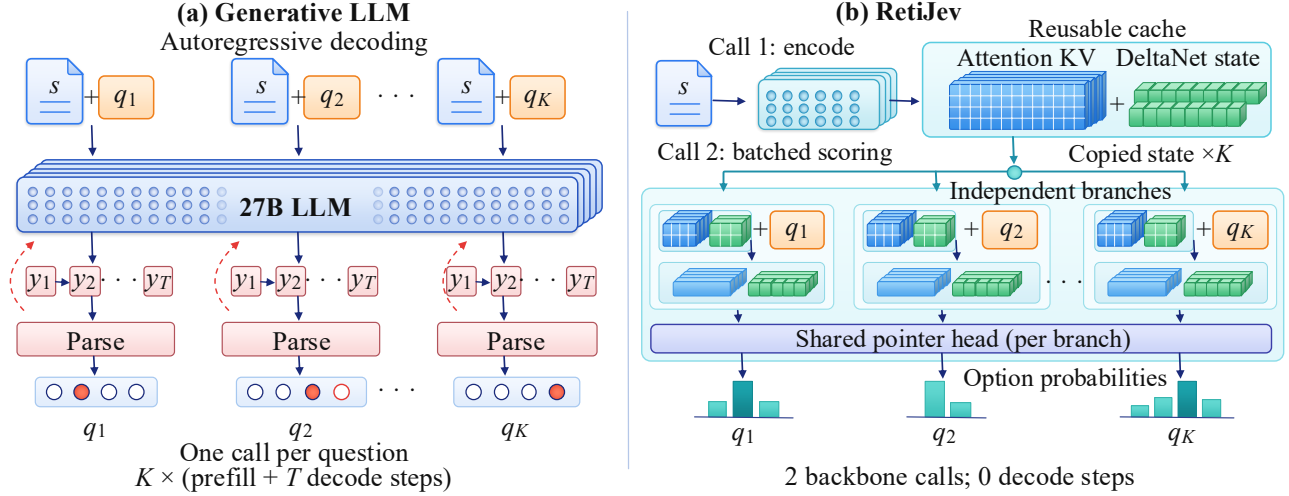}
  \caption{\textbf{Shared-context inference.}
(a) A generative LLM processes the patient state separately for each
question and autoregressively decodes an answer.
(b) RetiJev encodes the patient state once, replicates the cached attention
and recurrent states across independent question branches, and scores all
branches in parallel without autoregressive decoding.}
  \label{fig:inference-comparison}
\end{figure*}

\subsection{Training Objective}
\label{sec:method-objective}

\Cref{fig:arch}(c) summarizes the supervision used to jointly train the
model across the tasks in \cref{sec:data}. Their supervision has different structures: 
ordered ratings specify an ordering among answers, clinical decisions can
benefit from a teacher's relative preferences, and paraphrases should
preserve the answer to the same question.

\paragraph{Label supervision.}
Let $y_k\in\{1,\dots,m_k\}$ be the target candidate index and write
$p_{k,j}=p_\theta(o_{k,j}\mid s,q_k)$. Every training question contributes
the cross-entropy loss as
\begin{equation}
  \ell_{\mathrm{CE},k}=-\log p_{k,y_k}.
  \label{eq:ce}
\end{equation}
Following \medjev \citep{ma2026medjev}, we add the ranked probability score
(RPS) to cross-entropy for ordered ratings. With cumulative probabilities
$F_{k,r}=\sum_{j=1}^{r}p_{k,j}$, the ordinal term is
\begin{equation}
  \ell_{\mathrm{RPS},k}
  =\frac{1}{m_k-1}\sum_{r=1}^{m_k-1}
    \bigl(F_{k,r}-\mathbf{1}[y_k\le r]\bigr)^2.
  \label{eq:rps}
\end{equation}
This term applies to diagnosis-support ratings, urgency levels and ordinal
guideline grades.

\paragraph{Teacher distillation.}
For clinical-decision questions, the verified hard label remains the
primary training target. However, within a given candidate set, a one-hot
target does not encode the relative preference among the
choices. When available, we therefore augment the hard label with an
offline distribution from a frozen teacher (\qwenbig) \citep{qwen2026qwen38}. Let \(p^T_k\) denote the
teacher distribution over the candidates of question \(k\). We match the
temperature-softened teacher and student distributions using
knowledge distillation \citep{hinton2015distilling}:
\[
\ell_{\mathrm{KD},k}
=
\tau^2
\mathrm{KL}
\left(
p^T_{k,\tau}
\parallel
p_{\theta,k,\tau}
\right),
\]
where
\(p^T_{k,\tau}(j)\propto(p^T_k(j))^{1/\tau}\)
and \(p_{\theta,k,\tau}\) is obtained by applying temperature \(\tau\)
to the student logits. We apply this loss only when the teacher
distribution covers all candidates in the question. All teacher queries are
confined to training data, and paraphrased versions reuse the distribution
of their original case. We do not apply distillation to the guideline grading task,
whose targets are computed deterministically from the encoded criteria.

\paragraph{Diagnosis-coverage augmentation.}

Distillation on existing clinical-decision questions does not directly address the limited coverage of ophthalmic diseases. To broaden disease coverage, we construct additional single-best-diagnosis (diagnosis-choice) training cases using a vocabulary of 531 MeSH eye diseases \citep{lipscomb2000mesh}. For each differential diagnosis group, the LLM generates a synthetic patient description, identifies the most likely diagnosis, and provides a probability distribution over the candidate diagnoses. Only cases meeting the filtering criteria in \cref{app:generation} are retained. These cases are used exclusively for training.

\paragraph{Paraphrase consistency.}
Beyond covering more diseases, the model should produce stable predictions
when the same clinical findings are expressed in different words. For
clinical-decision training cases with paraphrase partners, we vary the
wording of the patient description while keeping the clinical findings,
question, candidate answers and target label fixed. We complement the
per-example label and distillation losses with an explicit constraint on
agreement between predictions across these descriptions.

For each case and question, we maintain an exponential moving average
$\bar p_k$ of \eyejev's own previous candidate distributions over the
original and paraphrased descriptions. The current prediction is encouraged
to match this detached target through
\begin{equation}
  \ell_{\mathrm{cons},k}
  =\operatorname{KL}\!\left(
    \operatorname{stopgrad}(\bar p_k)\,\|\,p_{\theta,k}\right),
  \label{eq:cons}
\end{equation}
where $p_{\theta,k}$ is \eyejev's candidate distribution for the description
currently being processed. The stop-gradient operation holds the target
fixed during the update, so gradients flow only through the current
prediction. Target updates and the conditions for applying this loss are
detailed in \cref{app:objectives}.

\paragraph{Joint optimization.}
Let $I_k^{\mathrm{ord}}$, $I_k^{\mathrm{KD}}$ and
$I_k^{\mathrm{cons}}$ be binary indicators for the applicability of the three auxiliary
terms. The per-question objective is
\begin{equation}
  \begin{aligned}
    \ell_k={}&\ell_{\mathrm{CE},k}
      +\lambda_{\mathrm{ord}} I_k^{\mathrm{ord}}\ell_{\mathrm{RPS},k}\\
      &+\lambda_{\mathrm{KD}} I_k^{\mathrm{KD}}\ell_{\mathrm{KD},k}
      +\lambda_{\mathrm{cons}} I_k^{\mathrm{cons}}\ell_{\mathrm{cons},k}.
  \end{aligned}
  \label{eq:question-loss}
\end{equation}
For each training case $c$ with $K_c$ questions, we average the question
losses before averaging over the training set $\mathcal D$. This gives
equal weight to cases regardless of their number of questions:
\begin{equation}
  \mathcal L(\theta)
  =\mathbb E_{c\sim\mathcal D}\!\left[
    \frac{1}{K_c}\sum_{k=1}^{K_c}\ell_{c,k}\right].
  \label{eq:training-loss}
\end{equation}
We use $\lambda_{\mathrm{ord}}=0.3$,
$\lambda_{\mathrm{KD}}=\lambda_{\mathrm{cons}}=0.5$ and $\tau=2$.
The base weights and token embeddings remain frozen, with only the LoRA adapters \citep{hu2022lora} and pointer head optimized. Training details are in \cref{tab:hparams}.
\newcommand{\both}{\textsuperscript{$\blacktriangle$}}
\newcommand{\one}{\textsuperscript{$\vartriangle$}}
\newcommand{\worse}{\textsuperscript{$\triangledown$}}
\begin{table*}[t]

\caption{\textbf{Accuracy on \bench} (\%).
  \emph{Unseen} denotes rule families excluded from training, whereas
  \emph{Seen} evaluates new patients from topics covered in training.
  Grading is evaluated on notes
  generated by an unseen renderer (OOD) and realistic threshold cases
  (Hard).
  Statistical markers indicate comparisons with the two baselines
  using paired bootstrap tests (95\% confidence):
  \both significantly better than both;
  \worse significantly worse than at least one;
  no marker indicates no significant difference from either.
  Latency is the median time per case (ms) on a single H200 GPU
  at batch size 1, with the 27B baseline evaluated using the letter-scoring protocol.}
  \label{tab:main}
  \centering
  \begin{small}
  \setlength{\tabcolsep}{4.5pt}
  \begin{tabular}{lcc cc cc cc}
    \toprule
    & & & \multicolumn{4}{c}{Clinical decisions (source-grounded labels)} & \multicolumn{2}{c}{Guideline grading (rule-derived labels)} \\
    \cmidrule(lr){4-7}\cmidrule(lr){8-9}
    & & & \multicolumn{2}{c}{Decision} & \multicolumn{2}{c}{Diagnosis} & \multicolumn{2}{c}{Grading} \\
    \cmidrule(lr){4-5}\cmidrule(lr){6-7}\cmidrule(lr){8-9}
    System & Params & Latency & Unseen & Seen & Unseen & Seen & OOD & Hard \\
    \emph{questions} & & & \emph{4{,}625} & \emph{3{,}106} & \emph{164} & \emph{224} & \emph{3{,}983} & \emph{4{,}230} \\
    \midrule
    \qwenbig, zero-shot            & 27B & 247 & 79.5 & 77.8 & \textbf{84.8} & 90.6 & 77.6 & 73.3 \\
    \jev, zero-shot (API)          & n/a & API & 79.5 & 77.4 & \textbf{84.8} & 90.2 & 83.0 & 79.4 \\
    \medjev, zero-shot             & 0.8B & 133 & 39.5 & 39.8 & 47.0 & 53.1 & 42.8 & 41.4 \\
    \midrule
    \eyejev-0.8B                   & 0.8B & 133 & 80.3 & 80.5\both & 68.9\worse & 86.2 & 95.9\both & 89.8\both \\
    \eyejev-2B                     & 2B & 136 & 81.4\both & 81.5\both & 72.6\worse & 87.1 & 96.1\both & 91.1\both \\
    \eyejev-9B                     & 9B & 158 & \textbf{84.9}\both & \textbf{83.9}\both & \textbf{84.8} & \textbf{91.5} & \textbf{97.3}\both & \textbf{96.3}\both \\
    \bottomrule
  \end{tabular}
  \end{small}
\end{table*}

\subsection{Shared-Context Inference}
\label{sec:method-inference}

When several questions concern the same patient, encoding the patient
state separately for each question repeats the same computation. As
illustrated in \cref{fig:inference-comparison}, we encode the state once
and score all questions in a single batch using its cached
representations. Each question is processed with only the patient state
and its own instruction and candidates, preserving the independence
defined in \cref{sec:method-model}.

The procedure requires two backbone calls per case: one for the shared
state and one for the batched question branches. If the state contains
$S$ tokens and question branch $k$ contains $B_k$ tokens, this reduces the
number of processed token positions from $KS+\sum_k B_k$ to
$S+\sum_k B_k$. The pointer head returns candidate probabilities without
autoregressive decoding. Cache reuse and branch isolation are detailed in
\cref{app:encoding}.

\section{Experiments}
\label{sec:exp}

\subsection{Setup}
\label{sec:exp-setup}

\paragraph{Baselines.}
\emph{Slow version:} \qwenbig \citep{qwen2026qwen38}, evaluated zero-shot under its most accuracy-favorable fast inference setting: letter scoring with reasoning disabled, requiring one forward pass per question \citep{robinson2023leveraging}.
\emph{Fast version:} \jev, a commercial general-purpose decision model served through an API \citep{jev2026}, also evaluated zero-shot, with one question submitted per request. All benchmark cases are synthetic, so no patient data were transmitted.
\emph{Same architecture without our data:} the released \medjev model \citep{ma2026medjev}, which uses the same model family as ours but is trained for clinical-variable extraction rather than on our decision data.

\paragraph{Evaluation.}
The primary metric is micro accuracy per test set, taking the
highest-probability option as the prediction. We estimate uncertainty
with 2{,}000 bootstrap resamples over cases and consider a pairwise difference
significant when the 95\% confidence interval of the paired difference
excludes zero \citep{efron1993introduction}. 

\subsection{Main Results}
\label{sec:exp-main}

Table~\ref{tab:main} reports results across three evaluation categories:
guideline grading, clinical decision, and diagnosis choice. We discuss these
categories separately to distinguish rule application, action selection, and
disease recognition.

\paragraph{Rule application transfers across note styles.}
On the eleven grading systems, every \eyejev model reaches
\accGradeOOD--\accGradeOODNineB when the notes are written by a model it never
saw (Grading-OOD) and \accGradeHard--\accGradeHardNineB on threshold cases
written as realistic notes with distractors (Grading-Hard). Both large zero-shot baselines are
significantly worse on both sets, by \gainGradeLow--\gainGradeHigh points. This measures the
benefit of targeted rule supervision over zero-shot use of general models.
These errors reflect failures to apply established clinical criteria despite explicit patient findings. For example, \jev incorrectly classifies diabetic retinopathy as severe when more than 20 haemorrhages are present in only three of the four retinal quadrants, and recommends AREDS2 supplementation for early AMD. The gains
persist under an unseen renderer and realistic threshold cases, supporting
rule application beyond familiar phrasing. That robustness depends on
paraphrase training: a model trained with one renderer deteriorated sharply
under wording changes (\cref{app:robust}).

\paragraph{Action selection improves with domain supervision and scale.}

On clinical decision-making tasks, \eyejev-9B significantly outperforms Qwen-27B and Jev 1.13 across Decision-Unseen (\accDecUNineB vs. 79.5\% for both baselines) and Decision-Seen (\accDecSNineB vs. 77.8\% and 77.4\%), with improvements of \gainDecNineLow--\gainDecNineHigh percentage points. The 2B variant also achieves significant gains over both baselines. Despite its smaller capacity, \eyejev-0.8B performs comparably to both baselines on Decision-Unseen and significantly outperforms them on Decision-Seen. Task-level results reveal the strengths of the 0.8B model: on unseen-source cases, it significantly outperforms both baselines in \emph{management direction} and exceeds Jev 1.13 in \emph{next investigation} and \emph{plan safety}, while remaining comparable to Qwen-27B (\cref{tab:tasks}). These findings demonstrate the effectiveness of \eyejev for clinical action selection across model scales.

\paragraph{Disease recognition remains a constraint for small models.}

On Diagnosis-Unseen, \eyejev-0.8B reaches \accDXu, compared with
\accDXuLLM for both large zero-shot baselines, and the 2B model also
remains below them. On Diagnosis-Seen, neither small model differs
significantly from either baseline. Scaling to 9B closes this gap:
\eyejev-9B reaches \accDXuNineB on Diagnosis-Unseen, with no significant
difference from either baseline on either diagnosis set. Because both
diagnosis sets are small, these comparisons have limited precision. In contrast to grading and action selection,
this pattern suggests that disease recognition is more sensitive to model
scale and the breadth of disease supervision ~\citep{madrid2026jev}.

\subsection{Speed}
\label{sec:exp-speed}

We timed every system from raw text to option probabilities at batch size 1
on development cases, all on one full H200 GPU (\cref{tab:speed},
\cref{app:latency}). \eyejev-0.8B answers a case, 2.65
questions on average, in \latOursHtwo with \memOurs of GPU memory, and in
\latOursCPU on eight CPU cores. Its latency is fixed per-call overhead rather
than computation, so the 2B and 9B backbones add only 3 and 25\,ms (\latNineB
for \eyejev-9B). \qwenbig on
the same GPU takes \latLLMScore per case with letter scoring, which decodes
nothing.
Generating the answer, as LLMs are usually deployed, costs \latLLMGen,
\speedupGenNineB{} times \eyejev-9B, and answering after reasoning takes
\latLLMThink (\tokThink tokens per question), about \speedupThinkNineB{}
times \eyejev-9B. 
Measurements varying the number of questions per case are in
\cref{app:latency}.

\subsection{Calibration and Deferral}
\label{sec:exp-cascade}
\begin{table}[t]
  \caption{\textbf{Per-case latency and peak GPU memory.}
Latency is reported as the median and 95th percentile (ms) on development cases with batch size 1. Peak memory is reported in GB. The 27B baseline is \qwenbig.}
  \label{tab:speed}
  \centering
  \begin{small}
  \setlength{\tabcolsep}{3pt}
  \begin{tabular}{llrrr}
    \toprule
    System & Hardware & p50 & p95 & GB \\
    \midrule
    \eyejev-0.8B & H200 & 133 & 144 & 3.3 \\
    \eyejev-0.8B & CPU, 8 cores & 347 & 798 & 5.9 \\
    \eyejev-2B & H200 & 136 & 162 & 8.0 \\
    \eyejev-9B & H200 & 158 & 171 & 31.5 \\
    \midrule
    27B, letter scoring & H200 & 247 & 512 & 51.4 \\
    27B, short answer & H200 & 546 & 934 & 51.3 \\  
    27B, with reasoning & H200 & 74{,}489 & 211{,}559 & 51.3 \\  
    \bottomrule
  \end{tabular}
  \end{small}
  \\[2pt]\parbox{\linewidth}{\scriptsize All GPU rows use one full NVIDIA H200. CPU rows: bf16 on 8 cores (0.8B fp32: 597\,ms median). CPU memory is peak resident set size.}
\end{table}

Reliable deployment requires confidence estimates that can support selective
deferral. On Decision-Unseen, \eyejev-0.8B and \eyejev-9B achieve expected
calibration errors (ECE) of 0.033 and 0.016, respectively, compared with 0.033 and
0.027 for the two large zero-shot baselines.
Confidence also ranks errors effectively. The corresponding areas under the
risk--coverage curve are 0.072 and 0.045, compared with 0.079 and 0.080 for
the baselines. At 80\% coverage, \eyejev-0.8B and \eyejev-9B reach 87.8\%
and 92.1\% accuracy, while the deferred 20\% of questions contains 51\% and
58\% of their errors, respectively. These results indicate that \eyejev's
probabilities can support selective prediction, allowing high-confidence
decisions to be handled locally while concentrating uncertain cases for
escalation.

\subsection{Ablation of the Training Objective}
\label{sec:exp-analysis}

\begin{table}[t]
  \caption{\textbf{Cumulative ablation of the training objective} for \eyejev-0.8B.
  Starting from cross-entropy, successive rows add KD, paraphrase consistency
  (Cons.) and RPS; the last row is the full objective.
  Decision and Diagnosis denote unseen-page accuracy (\%); Grading
  denotes OOD accuracy (\%). ECE is on Decision-Unseen (lower is better).}
  \label{tab:ablation}
  \centering\small
  \setlength{\tabcolsep}{2.5pt}
  \begin{tabular}{lcccc}
    \toprule
    Objective & Decision & Diagnosis & Grading & ECE \\
    \midrule
    Cross-entropy only & 77.8 & 66.5 & 95.9 & 0.186 \\
    + KD & 79.7 & 65.9 & 95.4 & 0.045 \\
    + KD + Cons. & 80.3 & 68.9 & 95.9 & 0.036 \\
    + KD + Cons. + RPS & 80.3 & 68.9 & 95.9 & 0.033 \\
    \bottomrule
  \end{tabular}
\end{table}

\Cref{tab:ablation} examines the contribution of the auxiliary training
objectives in \eyejev-0.8B. Starting from cross-entropy supervision, we
incrementally add teacher distillation and paraphrase consistency
(\cref{sec:method-objective}).

\paragraph{Teacher distillation improves decision learning and calibration.}
Adding teacher distillation increases Decision accuracy from 77.8\% to
79.7\% and reduces ECE substantially from 0.186 to 0.045. Diagnosis and
Grading accuracy change only slightly. This indicates that the main benefit
of distillation lies in improving clinical-decision learning and producing better-calibrated candidate probabilities.

\paragraph{Paraphrase consistency further improves robustness.}
Adding paraphrase consistency raises Decision accuracy to 80.3\% and Diagnosis accuracy to 68.9\%, while Grading returns to 95.9\% and ECE decreases further to 0.036. These gains suggest that enforcing prediction consistency across alternative renderings helps the model rely more on the underlying clinical evidence than on surface wording. Together, teacher distillation and paraphrase consistency account for the main improvement over the cross-entropy baseline.

\section{Limitations}
\label{sec:discussion}

Decision labels are source-grounded and cross-verified, but not
clinician-adjudicated, and all patient states are synthetic. \eyejev has also
not been evaluated prospectively. Clinician-labelled, real-world evaluation
is therefore needed before clinical use.

\section{Conclusion}
\label{sec:conclusion}

We introduced \eyejev, a family of compact models that formulate ophthalmic
clinical decisions as candidate scoring rather than text generation.
Source-grounded supervision enables strong rule application and clinical
action selection, while disease recognition remains more sensitive to model
scale and disease coverage. Overall, direct decision scoring
provides an efficient alternative to generative inference for structured
clinical decisions.

\section*{Impact Statement}
This work explores whether clinical decision support can be made fast and
local enough for use during an ophthalmic encounter. Compact decision models
could reduce dependence on cloud inference, improve responsiveness, and
support access to structured ophthalmic knowledge in settings with limited
specialist availability or connectivity.

These benefits must be weighed against the risks of incorrect or overconfident
recommendations, automation bias, and distribution shift beyond the synthetic
cases and source material used here. We mitigate these risks by reporting uncertainty and deferral behavior, transparently documenting failure cases, and limiting our claims to source-grounded benchmark performance. Moreover, \eyejev is a research
prototype and should not be used for clinical decision making without
independent clinical validation.

\bibliography{references}
\bibliographystyle{icml2026}

\newpage
\appendix
\onecolumn
\section{Action Catalog}
\label{app:catalog}

\begin{table}[h]
\caption{\textbf{Clinical-decision tasks and question formats.}
Action catalogs define the available options, from which case-specific candidates are selected. Urgency uses an ordinal scale, while diagnosis involves both individual disease ratings and most-likely diagnosis selection. The three question formats are detailed below.}
  \label{tab:catalog}
  \centering
  \small
  \begin{tabular}{@{}p{0.17\linewidth}p{0.10\linewidth}p{\dimexpr0.73\linewidth-4\tabcolsep\relax}@{}}
    \toprule
    Task & \shortstack[l]{Question\\formats} & Candidates or rating levels \\
    \midrule
    Urgency level & \emph{score} & \emph{4 levels:} routine follow-up (weeks--months) $<$ expedited outpatient review (days) $<$ same-day urgent evaluation $<$ immediate emergency management \\
    Referral target & \emph{noul}, \emph{choice} & \emph{14 actions:} manage in current setting; general ophthalmology; cornea/external disease; glaucoma; retina/vitreous; uveitis; neuro-ophthalmology; oculoplastics/orbit; paediatric/strabismus; ocular oncology; cataract/anterior-segment surgeon; non-ophthalmic specialty; emergency department; escalate \\
    Next investigation & \emph{noul}, \emph{choice} & \emph{21 actions:} none needed; clarify history; comprehensive eye examination; visual acuity and refraction; visual field; OCT (macula/RNFL); fundus photography/autofluorescence; fluorescein/ICG angiography; ocular ultrasound; corneal imaging; gonioscopy; anterior-segment OCT/UBM; orbital or neuro-imaging; laboratory/serology; microbiology; genetic testing; biopsy; electrophysiology; systemic/multidisciplinary evaluation; review records or repeat test; escalate \\
    Management direction & \emph{noul}, \emph{choice} & \emph{12 actions:} observation only; continue current management; adjust non-surgical management; start topical therapy; start systemic therapy; start intravitreal/periocular therapy; laser or office procedure; individualized surgical evaluation; stop or taper therapy; complete assessment before deciding; escalate to specialist/MDT; escalate \\
    Plan safety & \emph{noul}, \emph{choice} & \emph{6 actions:} proceed as planned; proceed with modification or precaution; defer until prerequisite completed; contraindicated, choose alternative; specialist decision; escalate \\
    Diagnosis & \emph{score}, \emph{choice} & \emph{Rating (3 levels):} not supported or excluded $<$ possible, needs differentiation $<$ leading working diagnosis. \emph{Selection:} the most likely diagnosis among case-specific disease candidates. \\
    \bottomrule
  \end{tabular}
\end{table}

\paragraph{Question formulation.}
Binary acceptability questions (\noul) assess an individual action with
answers no and yes.
Preferred and acceptable actions receive yes, inappropriate actions receive
no, and unresolved actions are not queried. For action selection, a \choice
question asks for the single best action only when exactly one is preferred. A \score question asks
for an ordered urgency level or rates each free-text diagnosis candidate from
excluded to leading. Verified per-candidate ratings can also yield a
most-likely-diagnosis \choice question when exactly one candidate is labeled
as the leading diagnosis. Each action-choice question shows
only the candidates relevant to its case, a median of six, rather than the
entire task catalog.

\section{Guideline Grading Systems}
\label{app:scales}

The benchmark covers ICDR diabetic retinopathy and macular oedema
\citep{wilkinson2003proposed}, the Beckman AMD classification with AREDS2
eligibility \citep{ferris2013clinical}, the clinical activity score for thyroid
eye disease \citep{mourits1997clinical}, ETROP treatment criteria for
retinopathy of prematurity \citep{etrop2003revised}, EVS initial management of
endophthalmitis \citep{evs1995results}, the Chandler classification of orbital
cellulitis \citep{chandler1970pathogenesis}, WHO visual-impairment categories
\citep{who2019world}, SUN anterior-chamber cell grades and activity change
\citep{jabs2005sun}, hyphema grade, Hodapp--Parrish--Anderson visual-field
staging \citep{hodapp1993clinical}, and whether an initial target intraocular
pressure has been reached \citep{gedde2021poag}.
\Cref{tab:scales} summarizes the clinical parameters, question types and
test-question counts for each system.

\begin{table}[h]
  \caption{\textbf{Grading systems and guideline rules.}
Labels are determined from the listed clinical information using predefined rules. Test-question counts are reported separately for standard, paraphrased, out-of-distribution, and hard cases.}
  \label{tab:scales}
  \centering
  \small
  \begin{tabular}{p{0.19\linewidth}p{0.28\linewidth}p{0.25\linewidth}r}
    \toprule
    System & Basis for labeling & Questions (type, levels) & Question numbers \\
    \midrule
    ICDR diabetic retinopathy & per eye: microaneurysms, quadrant haemorrhage counts, exudates, cotton-wool spots, venous beading, IRMA, new vessels, vitreous/preretinal bleeding, macular oedema and centre involvement & DR severity OD/OS (\score, 5); centre-involving DME OD/OS (\noul); referable DR (\noul) & 910 / 905 / 880 / 1{,}005 \\
    Beckman AMD & largest druse within 2 disc diameters of the fovea, pigment changes, geographic atrophy, new vessels & stage OD/OS (\score, 5); AREDS2 supplement indicated (\noul) & 621 / 618 / 543 / 636 \\
    Clinical activity score & seven inflammatory activity items in the assessed eye & CAS total (\score, 8); active TED, CAS $\ge 3$ (\noul) & 376 / 380 / 370 / 434 \\
    ROP (ETROP) & zone, stage, plus-disease status (absent, pre-plus, or plus) & ETROP type OD/OS (\choice, 3); treat now (\noul) & 627 / 627 / 627 / 642 \\
    Endophthalmitis (EVS) & precipitating procedure/event, elapsed time, presenting visual acuity & initial management (\choice, 3) & 201 / 201 / 196 / 220 \\
    Chandler & orbital imaging, cavernous-sinus thrombosis, proptosis, motility, acuity & group (\score, 5) & 204 / 204 / 190 / 91 \\
    WHO visual impairment & better-eye presenting visual acuity and central visual-field radius & category (\score, 7) & 207 / 207 / 185 / 214 \\
    SUN anterior chamber cells & anterior-chamber cell counts at current and previous visits & grade (\score, 6); activity change (\choice, 3) & 400 / 400 / 388 / 348 \\
    Hyphema & anterior-chamber blood level & grade (\score, 5) & 193 / 193 / 188 / 209 \\
    HPA visual-field staging & mean deviation, pattern-deviation counts, central sensitivity, hemifields & stage (\score, 3) & 204 / 204 / 204 / 218 \\
    Initial target IOP & current IOP relative to the benchmark target (75\% of untreated baseline) & at target (\noul) & 224 / 224 / 212 / 213 \\
    \bottomrule
  \end{tabular}
\end{table}

\paragraph{Case sampling and rendering.}
For each grading system, we sample 1{,}000 structured cases from predefined,
clinically plausible ranges, emphasizing values near decision thresholds,
such as 20 versus 21 haemorrhages per quadrant and drusen of 63 versus
64\,\textmu m. Labels are computed deterministically from the sampled
parameters using the corresponding grading rules. DeepSeek-V4.1-Flash
converts these cases into clinical notes using either a standard prompt,
with one of six documentation styles, or a paraphrase prompt, with two
case-specific rewriting directives, such as spelling out numbers or
describing the left eye first. Both prompts require preservation of the
clinical findings and measurements while omitting the computed grades,
stages and decisions. Training includes both standard and paraphrased notes.

\paragraph{Parameter-recovery verification.}
For standard and paraphrased notes, a separate call to the same
DeepSeek-V4.1-Flash model extracts structured parameters from the generated
text. This call receives only the note and the extraction schema, without
the original case parameters, computed labels or generation history.
Code compares the extracted values with the original parameters, and a
note is retained only if all required fields match exactly. Failed drafts
are regenerated and rechecked, with up to three generation attempts per
case in each rendering run. Standard-note acceptance rates range from
98.7 to 100\% across generation batches.

\paragraph{Test-set rendering.}
Grading-Standard and Grading-Paraphrase use held-out cases rendered with
the same model and respective prompts used for training, providing
in-distribution references (\cref{app:grading}). 
Grading-OOD uses the
paraphrase prompt with Qwen3.8-Flash.
Grading-Hard uses cases selected
for proximity to decision thresholds, rendered by Qwen3.8-Flash with clinical
shorthand and clearly dated information from earlier visits. For both sets, Qwen3.8-Flash is also used to extract grading parameters. Both OOD and Hard notes
must satisfy the same exact parameter-recovery criterion.

\begin{center}
\begin{minipage}{\linewidth}
  \captionsetup{type=table}
  \caption{\textbf{Ophthalmic content categories.}
  Categories may overlap; Miscellaneous groups topics outside the named categories.}
  \label{tab:ophthalmic-categories}
  \centering\small
  \begin{tabular}{@{}r@{\hspace{1em}}p{0.88\linewidth}@{}}
    \toprule
    No. & Category \\
    \midrule
    1 & Cataract / Anterior Segment \\
    2 & Cornea / External Disease \\
    3 & Glaucoma \\
    4 & Neuro-ophthalmology / Orbit \\
    5 & Oculoplastics / Orbit \\
    6 & Retina / Vitreous \\
    7 & Pediatric Ophthalmology / Strabismus \\
    8 & Uveitis \\
    9 & Refractive Management / Intervention \\
    10 & Ocular Trauma \\
    11 & Oncology / Pathology \\
    12 & Miscellaneous \\
    \bottomrule
  \end{tabular}
\end{minipage}
\end{center}

\section{Data Generation and Label Verification}
\label{app:generation}

\subsection{Source coverage and rule extraction}
\label{app:rule-inventory}
\label{app:rule-split}
We collected ophthalmology-related articles indexed in PubMed
and removed duplicate records, yielding \nPMCFullTexts{}
articles with accessible full texts.
We then applied keyword-based screening to identify articles
containing evidence relevant to diagnosis, investigation,
management, safety, referral, and urgency.
Relevant passages were extracted together with their
surrounding context and associated tables.
This process excluded 98.3\% of the collected articles,
retaining approximately 65{,}759 candidate articles for
subsequent rule extraction.

These passages are processed by DeepSeek-V4.1-Flash to extract further
structured clinical decision rules. Each rule specifies the clinical task, applicability conditions,
prerequisites, exceptions, supported decision, and
recommendation strength, together with verbatim supporting
evidence and source identifiers for traceability.
Semantically equivalent clinical decision rules across articles
were consolidated into \nExtractedRuleRecords{} rule families,
which were partitioned into disjoint training (80\%),
development (10\%), and held-out (10\%) pools. The \emph{Seen} evaluation cases
are independently generated from rule families represented
in training, whereas \emph{Unseen} cases are generated from held-out
rule families.

\subsection{Clinical-decision generation and verification}
We use DeepSeek-V4.1-Flash and Qwen3.8-Flash for case generation and independent cross-family verification. Each rule is used to generate 2–3 cases when supported by the evidence. Patient descriptions are limited to 200 words, and each supporting excerpt is limited to 40 words. During generation, the models preserve the original prerequisites, exceptions, and recommendation strength without adding unsupported thresholds, drug doses, or time windows. For verification, a model from the other family independently answers each question based on the case and source evidence, without access to the generated label. This step checks whether the answer is supported by the evidence and whether the rule applies to the case. Questions with matching answers pass the initial verification. Disagreements are further reviewed by Gemini-3.8-Flash and GPT-5.4-mini, with answer options presented in random order. These questions are retained only when both judges agree. All retained questions must have at least one direct supporting quote from the source and no insufficient-evidence judgment. Unresolved questions are excluded rather than treated as negative examples.

\subsection{External-disease distillation}
To broaden diagnostic coverage, we use an external
vocabulary of \nExtDiseases MeSH eye diseases selected independently of
the test sets. The \qwenbig teacher generates vignettes of 40--150 words
and supplies probability distributions over the target disease and
differential diagnoses drawn from MeSH siblings or three teacher-selected
alternatives. Retention requires agreement between the teacher's prediction
and the intended diagnosis, no explicit candidate names, and word-level
Jaccard similarity $\leq 0.6$ to every development or test case. This yields
\nExtCases synthetic cases, supplemented by 3{,}650 source-derived
diagnosis-choice training cases with teacher supervision. Development and
test texts are withheld from the teacher.

\section{Latency Measurement Protocol}
\label{app:latency}

\paragraph{Measurement scope.}
We measure end-to-end inference latency per case under a common protocol,
including input validation, tokenization and candidate scoring or answer
generation. Cases are
processed individually (batch size 1), and model loading is excluded.
GPU operations are synchronized immediately before and after each
measurement, and the first five cases serve as warm-up. The standard
sample comprises 300 development cases with 2.65 questions per case on
average (sampling seed 0). CPU measurements use 100 cases for the 0.8B
model. All latency measurements use development data.

\paragraph{Computational environment.}
GPU measurements use a single NVIDIA H200 card; CPU measurements use eight
host cores. The GPU software environment comprises PyTorch 2.14 with
CUDA 13.0, Transformers 5.17, flash-linear-attention 0.5.2 and causal-conv1d
1.7.0. CPU inference uses the reference PyTorch implementations of the
DeltaNet and convolution layers. Memory usage is measured as peak allocated
GPU memory or peak resident memory on CPU.

\paragraph{Baseline inference settings.}
For \qwenbig, we evaluate both candidate scoring and answer generation.
Candidate scoring uses the same letter-scoring protocol as the accuracy
evaluation. Greedy answer generation uses the model's chat
template, with a limit of 64 new tokens without reasoning (50 cases) or
2{,}048 new tokens with reasoning (20 cases).

\paragraph{Inference components and throughput.}
Profiling \eyejev-0.8B on 200 development cases on the same H200 attributes
0.9\,ms to input validation and tokenization, 29\,ms to state encoding,
1.9\,ms to cache replication, and 63\,ms to question processing and the
prediction head. A separate throughput experiment excludes preprocessing
and uses pre-encoded, independent state--question sequences, recomputing
the state for each question. This yields 19, 105 and 178 cases per second
at batch sizes of 1, 8 and 32 cases, respectively.

\begin{figure}[htbp]
  \centering
  \begin{minipage}{0.6\linewidth}
\begin{tikzpicture}
\begin{axis}[
  width=\linewidth, height=3.6cm,
  xmode=log, ymode=log, log basis x=10, unbounded coords=discard, 
  xtick={1,3,5,10,20}, xticklabels={1,3,5,10,20},
  xlabel={questions per case}, ylabel={ms per case (median)},
  xlabel style={font=\scriptsize}, ylabel style={font=\scriptsize},
  ticklabel style={font=\scriptsize},
  legend style={font=\scriptsize, at={(0.5,1.05)}, anchor=south,
    legend columns=2, draw=none, fill=none},
  grid=major, grid style={gray!20},
]
\addplot[mark=*, thick, blue!70!black] table[x=K, y=eyejev] {figures/data/speed_k.dat};
\addplot[mark=diamond*, thick, blue!40!cyan] table[x=K, y=eyejev9b] {figures/data/speed_k.dat};
\addplot[mark=square*, thick, orange!80!black] table[x=K, y=qwen08_per_question] {figures/data/speed_k.dat};
\addplot[mark=triangle*, thick, red!70!black] table[x=K, y=qwen27b_per_question] {figures/data/speed_k.dat};
\legend{{\eyejev-0.8B, state shared}, {\eyejev-9B, state shared}, {0.8B, one pass per question}, {27B, one pass per question}}
\end{axis}
\end{tikzpicture}
  \end{minipage}
  \caption{\textbf{Shared state encoding limits latency growth as the number
  of questions increases.} Median ms per case on one H200 at batch
  size 1, with each case's questions repeated to exactly $K$. \eyejev encodes
  the state once and scores all branches together, while the baselines score
  each question separately.}
  \label{fig:speed-k}
\end{figure}
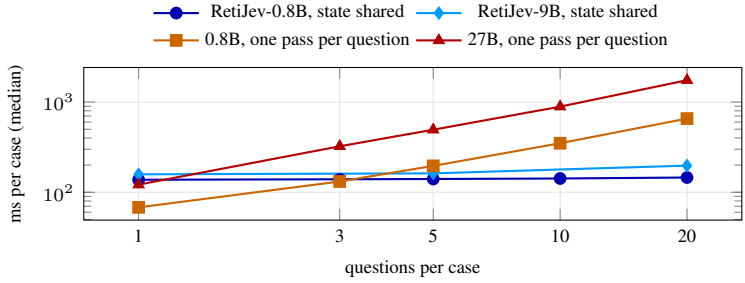

\paragraph{Scaling with questions per case.}
To isolate the cost of answering additional questions, we repeat each
case's questions to obtain exactly $K$ questions while holding its clinical
content fixed. The branches remain conditionally independent given the
shared state. As $K$ increases from one to twenty, \eyejev-0.8B latency
increases from \latOursKone to \latOursKtwenty, and \eyejev-9B reaches
\latNineBKtwenty (\cref{fig:speed-k}). In contrast, processing questions independently incurs approximately linear latency growth, reaching \latSmallKtwenty for the 0.8B backbone and \latLLMKtwenty for \qwenbig. Although a single backbone pass is faster for one or two questions, shared state encoding becomes increasingly efficient as the number of questions grows.

\section{Test Set Details}
\label{app:bench}

\begin{table}[h]
  \caption{\textbf{Composition of the \bench test sets.}
  The six main test sets contain \nMainTestQ questions; the two grading
  reference sets contain \nReferenceTestQ. Each grading set covers all
  eleven systems.}
  \label{tab:tiers}
  \centering
  \begin{small}
  \setlength{\tabcolsep}{3pt}
  \begin{tabular}{llr}
    \toprule
    Test set & What it tests & Questions \\
    \midrule
    \multicolumn{3}{l}{\emph{Main evaluation sets}} \\
    Decision-Unseen    & decisions, sources never used in training         & 4{,}625 \\
    Decision-Seen      & decisions, seen sources, new patients             & 3{,}106 \\
    Diagnosis-Unseen   & most likely diagnosis, unseen sources             & 164 \\
    Diagnosis-Seen     & most likely diagnosis, seen sources, new patients & 224 \\
    Grading-OOD        & 11 systems, renderer never used in training     & 3{,}983 \\
    Grading-Hard       & 11 systems, threshold cases, realistic notes    & 4{,}230 \\
    \midrule
    \multicolumn{3}{l}{\emph{Grading reference sets}} \\
    Grading-Standard   & 11 systems, training renderer (reference)       & 4{,}167 \\
    Grading-Paraphrase & 11 systems, training paraphraser (reference)    & 4{,}163 \\
    \midrule
                       & current benchmark total                        & \nTestQ \\
    \bottomrule
  \end{tabular}
  \end{small}
\end{table}

\paragraph{Clinical-decision and diagnosis evaluations.}
Decision-Seen and Decision-Unseen evaluate the six clinical-decision tasks,
including diagnosis-support ratings. Decision-Seen comprises new synthetic
cases from training rules, with near-duplicates of training
cases excluded; Decision-Unseen uses held-out sets. 
Diagnosis-Seen and Diagnosis-Unseen are derived from the corresponding
decision sets by retaining cases with verified ratings for all candidate
diagnoses and a unique leading diagnosis. Each retained case yields one
single-best-diagnosis question. Candidate order is randomized to control
positional bias: before randomization, 60\% of correct answers occupied
the first position.

\paragraph{Guideline grading.}
Grading-OOD evaluates robustness to notes generated by an LLM family not
used to produce the training notes. Grading-Hard combines cases near
decision thresholds with clinical shorthand and clearly dated findings
from earlier visits. Both use grading criteria covered during training.
Grading-Standard and Grading-Paraphrase provide reference evaluations using
the training generator and corresponding prompts. Generation configurations
and detailed results are provided in \cref{app:scales,app:grading}.

\section{Guideline Grading in Detail}
\label{app:grading}

\begin{table}[h]
  \caption{\textbf{Guideline grading in detail} (accuracy, \%). Top:
  in-distribution reference sets using the training renderers. Bottom:
  Grading-Hard results by question type. }
  \label{tab:grading}
  \centering
  \begin{small}
  \setlength{\tabcolsep}{4pt}
  \begin{tabular}{lrcccccc}
    \toprule
    & $n$ & 27B & \jev & \medjev & 0.8B & 2B & 9B \\
    \midrule
    Grading-Standard & 4{,}167 & 80.4 & 85.0 & 43.6 & 99.9 & 99.9 & 99.9 \\
    Grading-Paraphrase & 4{,}163 & 78.3 & 83.5 & 42.8 & 99.1 & 98.8 & 99.5 \\
    \midrule
    \multicolumn{8}{l}{\emph{Grading-Hard by question}} \\
    \quad AMD stage & 424 & 79 & 98 & 38 & 96 & 97 & 98 \\
    \quad AREDS2 indicated & 212 & 82 & 73 & 58 & 91 & 96 & 98 \\
    \quad CAS total & 217 & 75 & 88 & 0 & 82 & 84 & 96 \\
    \quad Chandler group & 91 & 99 & 99 & 53 & 95 & 98 & 99 \\
    \quad centre-involving DME & 402 & 100 & 100 & 83 & 98 & 95 & 100 \\
    \quad DR severity & 402 & 71 & 82 & 13 & 86 & 88 & 95 \\
    \quad EVS management & 220 & 66 & 78 & 35 & 90 & 81 & 100 \\
    \quad HPA stage & 218 & 57 & 72 & 36 & 86 & 87 & 89 \\
    \quad hyphema grade & 209 & 71 & 70 & 61 & 99 & 100 & 100 \\
    \quad target IOP reached & 213 & 57 & 54 & 54 & 80 & 79 & 90 \\
    \quad referable DR & 201 & 95 & 99 & 0 & 100 & 100 & 100 \\
    \quad ETROP type & 428 & 57 & 59 & 40 & 90 & 92 & 97 \\
    \quad ROP treat now & 214 & 73 & 75 & 54 & 92 & 96 & 99 \\
    \quad SUN cell grade & 174 & 98 & 99 & 66 & 99 & 100 & 100 \\
    \quad SUN activity change & 174 & 69 & 66 & 41 & 81 & 85 & 89 \\
    \quad active TED & 217 & 61 & 76 & 50 & 88 & 89 & 94 \\
    \quad WHO category & 214 & 47 & 56 & 26 & 70 & 79 & 91 \\
    \bottomrule
  \end{tabular}
  \end{small}
\end{table}

Grading-Standard and Grading-Paraphrase assess performance under the
training renderers, while Grading-Hard distinguishes performance on cases
near decision thresholds (\cref{tab:grading}). Every \eyejev model achieves
near-perfect accuracy on standard renderings (\accGradeStd). On
Grading-Hard, the lowest accuracies for the 0.8B model occur on WHO visual
impairment, target intraocular pressure, SUN activity change and the total
clinical activity score. These questions require comparing measurements
or combining findings across signs or visits. The 9B model achieves at
least 89\% accuracy on every question type. For \jev, the lowest accuracies
occur on ETROP type, WHO visual impairment and target intraocular pressure.

\section{Data Quality Criteria}
\label{app:filters}

Clinical-decision instances are screened for source fidelity, label
consistency and answer leakage. An instance is rejected if (i) a rule quote is not
verbatim on the page after repairing near-verbatim quotes (similarity
$\ge 0.9$); (ii) a candidate is outside the catalog, or the four label sets
overlap or do not cover the candidates; (iii) the state leaks the answer
(action names, recommendation phrases such as ``should'' or ``the next step
is'', or the disease name in a diagnosis state); (iv) the text is not English
or the state is outside 20--200 words; (v) a \choice instance has no contrast,
that is, neither a preferred nor an inappropriate option; (vi) an urgency
instance has more than one level; or (vii) a diagnosis labelled ``leading''
lacks a verbatim supporting phrase from the state. A ``possible'' diagnosis
without such evidence is relabelled as ``not supported or excluded''. Soft
recommendation language (``may'', ``can be considered'') is preserved when
assigning preferred and acceptable labels through explicit labelling rules.

\section{Robustness to Wording}
\label{app:robust}

\paragraph{Effect of paraphrase augmentation.}
We compare two 0.8B models in a grading pilot covering diabetic retinopathy,
AMD and the clinical activity score. One uses standard renderings only;
the other adds one paraphrase per training case while preserving the
underlying cases and labels. Evaluation uses standard notes, paraphrased
notes from the same generator, and paraphrased notes from an unseen
generator. The test cases are held out from training, and each condition
is restricted to questions answered by all four systems
(\cref{tab:wording-augmentation}).

\begin{table}[h]
  \caption{\textbf{Effect of paraphrase augmentation on wording robustness.}
  Accuracy (\%) in the grading pilot. Standard and paraphrase renderings
  use DeepSeek-V4.1-Flash; the unseen generator is Qwen3.8-Flash.
  The two 0.8B columns differ in training-text augmentation; 27B and \jev
  are zero-shot baselines. These pilot models are distinct from the final
  models in \cref{tab:main}.}
  \label{tab:wording-augmentation}
  \centering\small
  \begin{tabular}{lrcccc}
    \toprule
    & & \multicolumn{2}{c}{0.8B training renderings} & & \\
    \cmidrule(lr){3-4}
    Test rendering & Questions & Standard only & Standard + paraphrase & 27B & \jev \\
    \midrule
    Standard & 1{,}907 & 99.6 & 99.9 & 87.2 & 92.7 \\
    Paraphrase & 1{,}892 & 83.0 & 99.5 & 84.7 & 93.1 \\
    Unseen generator & 1{,}793 & 78.1 & 97.4 & 83.3 & 92.0 \\
    \bottomrule
  \end{tabular}
\end{table}

Standard-only training yields high accuracy on familiar wording but lower
accuracy than either baseline under both wording shifts. Paraphrase
augmentation improves accuracy under both shifts while preserving
in-distribution performance. All final \eyejev models use this
augmentation.

\section{Additional Analyses}
\label{app:analysis}

\begin{table}[h]
  \caption{\textbf{Decision-Unseen accuracy by task} (\%).
For management direction, 0.8B significantly outperforms both baselines, while for investigation and plan safety, it outperforms \jev and performs comparably to 27B. At 9B, improvements over both baselines are significant for management direction and plan safety, whereas investigation follows the same pattern as 0.8B (paired bootstrap, 95\% confidence intervals). Urgency and referral have insufficient sample sizes for reliable statistical comparisons.}
  \label{tab:tasks}
  \centering\small
  \setlength{\tabcolsep}{3pt}
  \begin{tabular}{lrccccc}
    \toprule
    & & \multicolumn{3}{c}{\eyejev} & & \\
    \cmidrule(lr){3-5}
    Task & $n$ & 0.8B & 2B & 9B & 27B & \jev \\
    \midrule
    Management direction & 1{,}925 & 83.8 & 85.0 & \textbf{86.9} & 79.9 & 81.7 \\
    Next investigation & 686 & 86.9 & \textbf{87.3} & \textbf{87.3} & 84.4 & 83.1 \\
    Plan safety & 997 & 79.6 & 80.5 & \textbf{86.1} & 78.9 & 74.6 \\
    Diagnosis (per cand.) & 961 & 69.5 & 70.8 & \textbf{78.0} & 75.3 & 77.4 \\
    Urgency & 50 & 74.0 & 78.0 & 84.0 & \textbf{86.0} & 84.0 \\
    Referral & 6 & \textbf{100.0} & \textbf{100.0} & \textbf{100.0} & \textbf{100.0} & \textbf{100.0} \\
    \bottomrule
  \end{tabular}
\end{table}

\paragraph{Adaptation settings.}
\Cref{tab:adaptation-settings} compares LoRA configurations and full
fine-tuning of the 0.8B backbone with the same training data. Rank 64 at a
learning rate of $3\times10^{-5}$ has the highest development accuracy
among these settings. Increasing adapter rank or updating all backbone
parameters does not by itself improve accuracy.  Configuration selection uses development data
only, and the selected 0.8B training configuration is applied unchanged
to the 2B and 9B models.

\begin{table}[h]
\caption{\textbf{Effects of LoRA configurations and fine-tuning strategies on the accuracy of the 0.8B backbone.}}
  \label{tab:adaptation-settings}
  \centering\small
  \begin{tabular}{lccr}
    \toprule
    Adaptation & LoRA rank & Learning rate & Accuracy (\%) \\
    \midrule
    LoRA & 16 & $10^{-4}$ & 86.4 \\
    LoRA & 64 & $10^{-4}$ & 85.1 \\
    LoRA & 64 & $3\times10^{-5}$ & 86.9 \\
    LoRA & 128 & $10^{-4}$ & -- \\
    Full fine-tuning & -- & $10^{-5}$ & 85.3 \\
    Full fine-tuning & -- & $5\times10^{-6}$ & 86.0 \\
    \bottomrule
  \end{tabular}
\end{table}

\paragraph{Generator-family bias.}
Because \qwenbig belongs to the Qwen family, questions written by Qwen models
could favour it. On the 2{,}221 DeepSeek-written Decision-Unseen questions,
\eyejev-0.8B reaches 85.0\%, against 82.3\% for 27B and 82.8\% for \jev.
On the 2{,}404 Qwen-written questions, the scores are closer: 76.0\% versus
77.0\% and 76.5\%. The 9B model reaches 88.6\% and 81.5\% on the two
respective groups. For these five systems, all scores are lower on the
Qwen-written group, which contains all 961 per-candidate diagnosis
questions. This task imbalance prevents attributing the difference to
model-family bias alone.

\section{Model and Training Details}
\label{app:hparams}

\subsection{Input Encoding and Branch Isolation}
\label{app:encoding}

Following \medjev, question $k$ is rendered as
\[
  \langle\textsc{s}\rangle\,s \;\Big\|\;
  \langle\textsc{q}\rangle\,x_k
  \big(\langle\textsc{o}\rangle\,o_{k,j}\,
  \langle/\textsc{o}\rangle\big)_{j=1}^{m_k}
  \langle\textsc{d}\rangle,
  \qquad k=1,\dots,K.
\]
The five delimiters are existing vocabulary tokens, with no new embeddings.
Question identifiers never appear in the text. Each branch restarts its
position indices immediately after the state, making its encoding independent
of the order in which questions are supplied.

For an attention-only backbone, a block-causal mask can let each branch
attend to the state and itself. The Qwen3.5 backbones interleave three Gated
DeltaNet layers with each gated full-attention layer. These layers carry
recurrent information between tokens, which an attention mask alone does
not isolate. The first backbone call caches the attention keys and
values together with the DeltaNet recurrent and convolution states. The
second call runs all branches as one batch using a separate copy of this
cache for each branch (\cref{fig:inference-comparison}). Together with the
branch-specific position indices, this gives each question the same causal
context as an independent state--question sequence.
On \bench, the median state length is \nStateTokMed tokens and the median
branch length is \nBranchTokMed, with \nQPerCase questions per case.

Let $h(\cdot)$ be a token's final hidden state. The pointer projections are
$u_k=W_Qh(\langle\textsc{d}\rangle_k)+b_Q$ and
$e_{k,j}=W_Kh(\langle/\textsc{o}\rangle_{k,j})+b_K$, with
$W_Q,W_K\in\mathbb{R}^{d_h\times d}$ and $d_h=256$.
The head supports between 2 and 255 options. Choice options are reshuffled
every training epoch.

\subsection{Training Objectives}
\label{app:objectives}

\paragraph{Supervision by question type.}
\cref{sec:method-objective} defines the training objective. Cross-entropy
applies to every question, including binary judgments represented as
two-candidate distributions. Ordinal supervision applies to \score
questions, including diagnosis-support ratings and
urgency levels as well as ordinal guideline grades. Distillation is enabled
only for clinical-decision questions with a teacher distribution containing
every current candidate. Candidate identities align the teacher
probabilities with the student logits, including after choice-option
permutation. Teacher probabilities are clamped below at $10^{-6}$, raised
to the power $1/\tau$ and normalized. Paraphrases reuse the teacher
distribution of their original case.

\paragraph{Consistency targets.}
For each case--question pair, consistency supervision uses a detached
moving average of previous predictions. The target is applied only when
the most recent prediction was obtained from a different rendering of the
same case and the candidate identities and order match the current
question. After computing the loss, the target is updated as
\[
  \bar p_k\leftarrow 0.5\,\bar p_k
    +0.5\,\operatorname{stopgrad}(p_{\theta,k}).
\]
If no target with matching candidates exists, the current detached
prediction initializes it. This update defines an exponential moving
average over successive predictions. Paraphrase consistency is restricted
to clinical-decision cases. Neither distillation nor consistency applies
to deterministically computed grading
labels. All active loss terms are averaged over questions within each case
and then over cases, as in \cref{eq:training-loss}.

\begin{table}[h]
  \caption{\textbf{Training configuration} shared by all \eyejev model sizes.}
  \label{tab:hparams}
  \centering
  \small
  \begin{tabular}{ll}
    \toprule
    Backbones & Qwen3.5-\{0.8B, 2B, 9B\}-Base, frozen; 0.75B / 1.88B / 8.95B text parameters \\
              & hybrid layouts: 18 DeltaNet $+$ 6 attention layers (0.8B, 2B), 24 $+$ 8 (9B) \\
    Adapters & LoRA rank 64, $\alpha=128$, dropout 0.05, on all attention, DeltaNet and MLP projections \\
    Head & pointer head, $d_h=256$, trained from scratch \\
    Trainable parameters & 43.8M / 68.3M / 175.2M \\
    Optimizer & AdamW, lr $3\times10^{-5}$, weight decay 0.01, one-cycle (5\% warm-up), clip 1.0 \\
    Batch, epochs & 4 cases per step, 3 epochs, bf16 autocast, gradient checkpointing \\
    Losses & cross-entropy $+$ 0.3 RPS (\score); distillation 0.5 ($\tau=2$, decisions only); consistency 0.5 \\
    Data & \nTrainRec records, \nTrainQ questions; \choice options reshuffled every epoch \\
    Hardware, time & one H200 GPU: 2.5 / 2.4 / 4.4 hours; peak memory 6.1 / 12.4 / 42.0\,GiB \\
    Selection & configuration selected on development data at 0.8B using a prespecified criterion; unchanged at 2B and 9B \\
    \bottomrule
  \end{tabular}
\end{table}

\subsection{Training Supervision}
\label{app:training-data}

The training set supports three complementary tasks
(\cref{tab:training-data}): choosing clinical actions and rating diagnostic
support, selecting the most likely diagnosis, and applying guideline
criteria. The first two tasks form the clinical-decision and diagnosis
component; the third forms the guideline-grading component.
Clinical-decision questions use source-derived
cases. Diagnosis selection uses 6{,}882 questions from these cases and
6{,}933 from synthetic cases covering \nExtDiseases external disease terms.
Guideline-grading labels are computed from the \nScales encoded systems and
rules described in \cref{app:scales}.

\begin{center}
\begin{minipage}{\linewidth}
  \captionsetup{type=table}
  \caption{\textbf{Supervision in the training set.}
  All three \eyejev model sizes use this same training set.
  Counts include the original and paraphrased examples used for training.}
  \label{tab:training-data}
  \centering\small
  \begin{tabular}{@{}p{0.20\linewidth}p{0.64\linewidth}r@{}}
    \toprule
    Task & Training target & Questions \\
    \midrule
    Clinical decisions & Action selection, action acceptability, urgency,
      and support ratings for individual candidate diagnoses. & 51{,}411 \\
    Diagnosis selection & The single most likely diagnosis among the candidate
      diseases for a case. & 13{,}815 \\
    Guideline grading & Grades and clinical recommendations determined by
      the \nScales encoded grading systems and rules. & 29{,}335 \\
    \midrule
    \textbf{Total} & & \textbf{\nTrainQ} \\
    \bottomrule
  \end{tabular}
\end{minipage}
\end{center}

\section{Examples of Every Task}
\label{app:example}

\paragraph{Clinical Decision.}

\exbox{%
\textbf{Urgency level.} \emph{Cherry-Red Spot}.\\
\textbf{State.} A 65-year-old man presents to the emergency department with sudden, painless loss of vision in his right eye that occurred 2 hours ago. He has no history of trauma, recent surgery, or known metabolic disorders. Examination reveals a cherry red spot at the macula of the right eye. The left eye is normal. The team must now decide on the level of urgency for the evaluation.\\
\textbf{Question} (\score): What level of clinical escalation does the current state require?\\
\textbf{Levels.} Routine follow-up or elective review (weeks to months) $<$ Expedited outpatient evaluation within days $<$ Same-day urgent ophthalmic evaluation $<$ Immediate emergency management; cannot wait for a clinic visit.\\
\textbf{Label.} Immediate emergency management; cannot wait for a clinic visit.\\
\textbf{Retained excerpt} (Management): ``Cherry red spot at the macula is a clinically significant fundoscopic finding that usually denotes an ocular emergency, especially in the patient who presents with sudden, painless unilateral vision loss and absence of other preceding clinical factors.''
}

\exbox{%
\textbf{Urgency level.} \emph{Dysthyroid Optic Neuropathy}.\\
\textbf{State.} A 45-year-old woman with active Graves orbitopathy and recently documented optic nerve dysfunction started intravenous methylprednisolone pulse therapy one week ago. Before treatment her acuity was 20/80 OD with a right afferent pupillary defect and an inferior arcuate field defect; today acuity is 20/30 OD with improved colour vision and a smaller field defect. She remains on weekly pulse therapy. She has mild proptosis, no diplopia, no pain and no new visual symptoms between visits. Blood glucose, blood pressure, liver enzymes and electrolytes are normal. She asks how her vision will be followed over the coming weeks while treatment continues.\\
\textbf{Question} (\score): What level of clinical escalation does the current state require?\\
\textbf{Levels.} Routine follow-up or elective review (weeks to months) $<$ Expedited outpatient evaluation within days $<$ Same-day urgent ophthalmic evaluation $<$ Immediate emergency management; cannot wait for a clinic visit.\\
\textbf{Label.} Expedited outpatient evaluation within days.\\
\textbf{Retained excerpt} (Management $>$ Orbital Decompression): ``Urgent orbital decompression is considered if there is no improvement of optic nerve function within 2 weeks following medical treatment.''
}

\exbox{%
\textbf{Referral target.} \emph{Ocular Manifestations of Isotretinoin}.\\
\textbf{State.} A 19-year-old woman has been taking isotretinoin 0.5 mg/kg daily for nodular acne for 6 weeks. She presents to her dermatology clinic reporting 1 week of gritty foreign-body sensation in both eyes that has not improved with over-the-counter artificial tears she has used 4 times daily. Visual acuity is unchanged from baseline, and there is no pain, redness, discharge, or vision loss. The dermatologist must now decide where to direct her care.\\
\textbf{Question} (\choice): Where, if anywhere, does this patient need to be referred at this point? Choose the referral direction.\\
\textbf{Options.} (1) Manage in the current setting; no referral needed. (2) Refer to general ophthalmology (from primary care, emergency or optometry settings). (3) Refer to cornea / external disease subspecialty. (4) Refer to a non-ophthalmic specialty (neurology, rheumatology, endocrinology, infectious disease, etc.). (5) Outside this catalog; escalate for a human decision.\\
\textbf{Labels.} preferred: \emph{Refer to general ophthalmology (from primary care, emergency or optometry settings)}. inappropriate: \emph{Manage in the current setting; no referral needed}.\\
\textbf{Also asked} (\noul): ``In the current state, is 'Manage in the current setting; no referral needed' an acceptable referral decision?'' $\rightarrow$ no; ``In the current state, is 'Refer to general ophthalmology (from primary care, emergency or optometry settings)' an acceptable referral decision?'' $\rightarrow$ yes.\\
\textbf{Retained excerpt} (Diagnosis $>$ Diagnosis and Workup): ``The dermatologist should refer the patient to an ophthalmologist if the patient is experiencing any new ocular pain, eyelid swelling, unresponsive dry eye, decreased vision, reduced night vision, visual field alterations, color perception alterations, persistent foreigh body sensation, eyelid abnormality, and/or conjunctival secretions.''
}

\exbox{%
\textbf{Referral target.} \emph{Shaken Baby Syndrome}.\\
\textbf{State.} A 6-month-old infant with a documented earlier episode of suspected abusive head trauma is seen in a pediatric ophthalmology clinic six weeks later. The intraretinal hemorrhages noted at presentation have largely resolved. A dense nonclearing vitreous hemorrhage persists in the right eye, and B-scan ultrasound shows a macula-sparing tractional retinal detachment in that eye. The left eye shows no detachment and no vitreous hemorrhage. No full-thickness macular hole is documented. Peripheral nonperfusion has not been assessed with fluorescein angiography. The infant is medically stable, feeding well, and is not in an emergency department. Pupils are reactive in both eyes. No surgical plan has been made.\\
\textbf{Question} (\choice): Where, if anywhere, does this patient need to be referred at this point? Choose the referral direction.\\
\textbf{Options.} (1) Refer to the emergency department / systemic emergency pathway. (2) Refer to a non-ophthalmic specialty (neurology, rheumatology, endocrinology, infectious disease, etc.). (3) Refer to retina / vitreous subspecialty. (4) Refer to general ophthalmology (from primary care, emergency or optometry settings). (5) Outside this catalog; escalate for a human decision.\\
\textbf{Labels.} preferred: \emph{Refer to retina / vitreous subspecialty}.\\
\textbf{Also asked} (\noul): ``In the current state, is 'Refer to retina / vitreous subspecialty' an acceptable referral decision?'' $\rightarrow$ yes.\\
\textbf{Retained excerpt} (Shaken Baby Syndrome or Non Accidental Trauma $>$ Management/Treatment): ``Surgical vitrectomy may be needed for nonclearing vitreous hemorrhage, macular hole, or retinal detachment, but procedures are complex in infant eyes and often require consultation with a pediatric retina trained surgeon.''
}

\exbox{%
\textbf{Next investigation.} \emph{Imaging in Ocular and Ocular Adnexal Trauma}.\\
\textbf{State.} A 29-year-old woman presents after a high-velocity metal injury to the left eye. She reports decreased vision and floaters. Visual acuity is hand motions in the left eye. There is a corneal opacity and a small hyphema, preventing a clear view of the posterior segment. The globe contour appears intact, and there is no obvious subconjunctival air or wound leak. The clinical concern includes a retained intraocular foreign body and associated posterior segment injury. The patient has no known metallic foreign body elsewhere. She is systemically stable.\\
\textbf{Question} (\choice): What is the most appropriate next investigation for the decision at hand? Choose one.\\
\textbf{Options.} (1) Orbital or neuro-imaging (CT or MRI). (2) Ocular ultrasound (B-scan / A-scan). (3) No additional investigation; current information suffices for this decision. (4) Outside this catalog; escalate for a human decision.\\
\textbf{Labels.} preferred: \emph{Orbital or neuro-imaging (CT or MRI)}. also acceptable: \emph{Ocular ultrasound (B-scan / A-scan)}.\\
\textbf{Also asked} (\noul): ``In the current state, is 'Orbital or neuro-imaging (CT or MRI)' an acceptable next investigation?'' $\rightarrow$ yes; ``In the current state, is 'Ocular ultrasound (B-scan / A-scan)' an acceptable next investigation?'' $\rightarrow$ yes.\\
\textbf{Retained excerpt} (Imaging $>$ Computerized tomography (CT scan)): ``Computerized tomography is the imaging of choice for evaluating suspected orbital and orbitofacial fractures, intraorbital and intraocular foreign bodies and traumatic optic neuropathies.''\\
\textbf{Retained excerpt} (Imaging $>$ Ultrasonography): ``It is useful in detecting intraocular foreign bodies (IOFB) and outlining soft tissue abnormalities of the eye and orbit in presence of media opacity.''
}

\exbox{%
\textbf{Next investigation.} \emph{Retinal Pigment Epithelial Tears}.\\
\textbf{State.} A 76-year-old man with neovascular age-related macular degeneration in the left eye, treated with intravitreal anti-VEGF injections every 6 weeks, presents two days after his latest injection with sudden severe loss of central vision. Visual acuity is 20/400 OS. Dilated fundus examination shows a crescent-shaped, well-defined hyperpigmented line in the macula with an adjacent circumscribed depigmented area of bare choroid, together with scattered subretinal hemorrhages. Intraocular pressure is 15 mmHg. There is no pain and no history of trauma. No retinal imaging has been obtained at today's visit and no previous imaging is available; the clinical picture is thought to represent a retinal pigment epithelial tear.\\
\textbf{Question} (\choice): What is the most appropriate next investigation for the decision at hand? Choose one.\\
\textbf{Options.} (1) OCT of the macula and/or optic nerve head RNFL. (2) Fluorescein and/or ICG angiography. (3) Fundus photography and/or fundus autofluorescence. (4) No additional investigation; current information suffices for this decision. (5) Clarify history, medications, family history or symptom details. (6) Outside this catalog; escalate for a human decision.\\
\textbf{Labels.} preferred: \emph{OCT of the macula and/or optic nerve head RNFL}. also acceptable: \emph{Fluorescein and/or ICG angiography}, \emph{Fundus photography and/or fundus autofluorescence}.\\
\textbf{Also asked} (\noul): ``In the current state, is 'OCT of the macula and/or optic nerve head RNFL' an acceptable next investigation?'' $\rightarrow$ yes; ``In the current state, is 'Fluorescein and/or ICG angiography' an acceptable next investigation?'' $\rightarrow$ yes; and 1 more.\\
\textbf{Retained excerpt} (Diagnosis): ``The diagnosis of RPE tears is clinical, and can be assisted by different imaging modalities, with optical coherence tomography (OCT) imaging being one of the best ways to confirm an RPE tear.''
}

\exbox{%
\textbf{Management direction.} \emph{Congenital and Acquired Epiblepharon}.\\
\textbf{State.} A 6-year-old East Asian boy has a 2-year history of bilateral eye rubbing and tearing and has used topical lubricants regularly for the past 6 months with good adherence. Examination shows a medial lower eyelid skin fold with lashes touching the inferomedial cornea in down gaze and adduction; the eyelid margin and posterior lamella are in normal position. No fluorescein staining and no slit lamp documentation of the corneal surface have been performed, so the severity of any keratopathy is not known. His symptoms have persisted during the lubricant trial. There is no history of corneal scarring.\\
\textbf{Question} (\choice): Which management direction is preferred now? Choose the direction, not a specific procedure.\\
\textbf{Options.} (1) Observation and monitoring without starting treatment. (2) Start topical medical therapy. (3) Proceed to individualized surgical evaluation (procedure not specified). (4) Complete the assessment that determines the management choice before deciding. (5) Outside this catalog; escalate for a human decision.\\
\textbf{Labels.} preferred: \emph{Complete the assessment that determines the management choice before deciding}.\\
\textbf{Also asked} (\noul): ``In the current state, is 'Complete the assessment that determines the management choice before deciding' an acceptable management direction?'' $\rightarrow$ yes.\\
\textbf{Retained excerpt} (Management): ``Indications for surgical intervention include persistent symptoms, significant keratopathy despite conservative management or inability to frequently apply lubricants especially in young children..''
}

\exbox{%
\textbf{Management direction.} \emph{Diabetic Macular Ischemia}.\\
\textbf{State.} A 63-year-old female with 35 years of type 2 diabetes presents with decreased vision in both eyes, 20/80 OD, 20/50 OS. Examination shows pseudophakia with clear media. Fundus examination reveals a ``featureless retina'' with minimal hemorrhages and microaneurysms, along with a poor foveal reflex. There is a chronic neovascular frond stable for multiple months. She had extensive panretinal photocoagulation. Central involved macular edema is present on OCT.\\
\textbf{Question} (\choice): Which management direction is preferred now? Choose the direction, not a specific procedure.\\
\textbf{Options.} (1) Start intravitreal or periocular injection therapy. (2) Observation and monitoring without starting treatment. (3) Start systemic medical therapy. (4) Outside this catalog; escalate for a human decision.\\
\textbf{Labels.} preferred: \emph{Start intravitreal or periocular injection therapy}. inappropriate: \emph{Observation and monitoring without starting treatment}.\\
\textbf{Also asked} (\noul): ``In the current state, is 'Start intravitreal or periocular injection therapy' an acceptable management direction?'' $\rightarrow$ yes.\\
\textbf{Retained excerpt} (Management $>$ Diabetic Macular Edema with Diabetic Macular Ischemia -- Controversies and Response to Therapy): ``Anti-VEGF therapy for DME is not a contraindication in the setting of DMI''
}

\exbox{%
\textbf{Plan safety.} \emph{Pre-Ophthalmologist Management of Eye Trauma}.\\
\textbf{State.} A 19-year-old man presents 6 hours after blunt trauma to the right eye. A 2 mm layered hyphema occupies the inferior anterior chamber and intraocular pressure is 16 mmHg. angle examination with a contact lens of the right eye shows a shallow anterior chamber with an occludable angle, and there is no history of a prior angle-closure attack. He has no coagulopathy, no sickle cell disease, and takes no anticoagulants. Visual acuity is 20/40 OD. He reports marked photophobia. There is no pain requiring analgesia. The plan is to start topical atropine 1\% drops and to discharge him with ophthalmology follow-up.\\
\textbf{Question} (\choice): Given the current state, what is the correct judgement about the proposed plan?\\
\textbf{Options.} (1) The plan can proceed as planned. (2) Can proceed, but with modification or added precautions. (3) Defer until a prerequisite is completed. (4) Contraindicated; choose an alternative. (5) Escalate for a specialist decision. (6) Outside this catalog; escalate for a human decision.\\
\textbf{Labels.} preferred: \emph{Contraindicated; choose an alternative}. inappropriate: \emph{The plan can proceed as planned}.\\
\textbf{Also asked} (\noul): ``In the current state, is 'The plan can proceed as planned' an acceptable judgement about the proposed plan?'' $\rightarrow$ no; ``In the current state, is 'Contraindicated; choose an alternative' an acceptable judgement about the proposed plan?'' $\rightarrow$ yes.\\
\textbf{Retained excerpt} (Pre-ophthalmologic Management of Threats to Vision $>$ Traumatic Hyphema): ``In the absence of narrow angle glaucoma, a mydriatic-cycloplegic such as atropine 1\% can attenuate photophobia and stabilize the blood aqueous barrier.''
}

\exbox{%
\textbf{Plan safety.} \emph{Preoperative Evaluation for LASIK Surgery}.\\
\textbf{State.} A 28-year-old man is scheduled for LASIK next week in a refractive surgery clinic. Refraction and visual acuity are stable at -4.25 D OU, and anterior chamber, iris, lens, and dilated fundus examination are normal. Meibomian gland evaluation shows lid disease with inspissated orifices. Tear break-up time is 4 seconds OU and Schirmer testing is low-normal; fluorescein and lissamine green staining show punctate keratitis inferiorly OU. Dryness treatment with lid hygiene and tear substitutes was started two weeks ago, and repeat testing today shows the same abnormal tear break-up time and unchanged staining. The surgeon must decide today whether the planned LASIK can go ahead.\\
\textbf{Question} (\choice): Given the current state, what is the correct judgement about the proposed plan?\\
\textbf{Options.} (1) The plan can proceed as planned. (2) Can proceed, but with modification or added precautions. (3) Defer until a prerequisite is completed. (4) Contraindicated; choose an alternative. (5) Escalate for a specialist decision. (6) Outside this catalog; escalate for a human decision.\\
\textbf{Labels.} preferred: \emph{Defer until a prerequisite is completed}. inappropriate: \emph{The plan can proceed as planned}.\\
\textbf{Also asked} (\noul): ``In the current state, is 'The plan can proceed as planned' an acceptable judgement about the proposed plan?'' $\rightarrow$ no; ``In the current state, is 'Defer until a prerequisite is completed' an acceptable judgement about the proposed plan?'' $\rightarrow$ yes.\\
\textbf{Retained excerpt} (LASIK Testing $>$ Dry Eye Testing): ``Patients will need continual follow up and re-evaluation until dryness is improved, before LASIK can be performed.''
}

\exbox{%
\textbf{Differential diagnosis.} \emph{Orbital Cellulitis}.\\
\textbf{State.} A 7-year-old girl presents with left eyelid redness and swelling for 1 day. She has no pain with eye movements, no proptosis, and full extraocular motility. Visual acuity is 20/20 OS with no relative afferent pupillary defect. She is afebrile and well-appearing. The team must now decide which diagnosis best explains the findings.\\
\textbf{Questions} (\score, one per candidate): ``Given the current state, how well is the diagnosis \emph{X} supported?'' Levels: not supported or excluded $<$ possible, needs differentiation $<$ leading working diagnosis.\\
\textbf{Labels.} Orbital cellulitis: \textbf{not supported}; Preseptal cellulitis: \textbf{leading} (evidence: ``left eyelid redness and swelling''); Idiopathic inflammation/specific inflammation: \textbf{not supported}; Neoplasia: \textbf{not supported}; Thyroid eye disease: \textbf{not supported}.\\
\textbf{Retained excerpt} (Diagnosis): ``The presence of the following signs is suggestive of orbital involvement: proptosis, chemosis, pain with eye movements, ophthalmoplegia''
}

\exbox{%
\textbf{Differential diagnosis.} \emph{Sorsby Macular Dystrophy}.\\
\textbf{State.} A 38-year-old male reports worsening central vision and distorted shapes. His mother experienced similar vision decline in her 30s. Slit lamp exam shows yellow deposits located between Bruch's membrane and the RPE. No honeycombing or egg-yolk lesions are observed. The team must now determine the leading inherited macular dystrophy diagnosis.\\
\textbf{Questions} (\score, one per candidate): ``Given the current state, how well is the diagnosis \emph{X} supported?'' Levels: not supported or excluded $<$ possible, needs differentiation $<$ leading working diagnosis.\\
\textbf{Labels.} Sorsby Macular Dystrophy: \textbf{leading} (evidence: ``yellow deposits located between Bruch's membrane and the RPE''); Malattia Leventinese: \textbf{not supported}; Pattern dystrophy: \textbf{possible} (evidence: ``worsening central vision and distorted shapes''); Best disease: \textbf{not supported}.\\
\textbf{Retained excerpt} (Diagnosis): ``symptoms of central visual blurring prior to the age of 40 years old may be an indication of SMD''\\
\textbf{Retained excerpt} (Diagnosis $>$ Physical Examination): ``yellow, extracellular deposits between Bruch's membrane and the RPE''
}

\paragraph{Guideline grading.}

\exbox{%
\textbf{ICDR diabetic retinopathy (case parameters).} OD: haemorrhages per quadrant (ST, SN, IT, IN) 12, 30, 24, 34; venous beading in 0 quadrants; IRMA quadrants 0; new vessels no; vitreous haemorrhage no; macular oedema non-central. OS: haemorrhages per quadrant (ST, SN, IT, IN) 13, 5, 18, 11; venous beading in 3 quadrants; IRMA quadrants 0; new vessels no; vitreous haemorrhage no; macular oedema central.\\
\textbf{Standard rendering.} 36-year-old man with type 2 diabetes of 24 years' duration, HbA1c 10.5\%, seen in the diabetic eye clinic. Visual acuity is 6/6 in the right eye and 6/6 in the left eye. On dilated fundus examination of the right eye, microaneurysms and hard exudates are present. Intraretinal haemorrhages by quadrant are superotemporal 12, superonasal 30, inferotemporal 24 and inferonasal 34. There are no cotton-wool spots, no venous beading, no intraretinal microvascular abnormalities, no new vessels at the disc or elsewhere, and no vitreous or preretinal haemorrhage. OCT of the right eye shows retinal thickening and intraretinal fluid outside the central 1 mm subfield, with the central subfield spared. In the left eye, microaneurysms and hard exudates are present, with 4 cotton-wool spots and definite venous beading in 3 quadrants. Intraretinal haemorrhages by quadrant are superotemporal 13, superonasal 5, inferotemporal 18 and inferonasal 11. There are no intraretinal microvascular abnormalities, no new vessels at the disc or elsewhere, and no vitreous or preretinal haemorrhage. OCT of the left eye shows retinal thickening and intraretinal fluid involving the central 1 mm subfield.\\
\textbf{Paraphrase rendering.} 36yo man, type 2 diabetes 24 yrs, HbA1c 10.5. VA 6/6 both eyes.\\ OD: MAs present. IRH in all four quadrants: upper temporal 12, upper nasal 30, lower temporal 24, lower nasal 34. HEx present. No CWS, no VB, no IRMA, no NVD/NVE, no VH. OCT: thickening/fluid outside the central 1 mm subfield.\\ OS: MAs present. IRH: upper temporal 13, upper nasal 5, lower temporal 18, lower nasal 11. HEx present. CWS 4. VB in 3 quadrants. No IRMA, no NVD/NVE, no VH. OCT: thickening/fluid involving the central 1 mm subfield.\\
\textbf{Labels.} \texttt{dr\_severity\_od}: Moderate NPDR: more than just microaneurysms but less than severe NPDR; \texttt{dr\_severity\_os}: Severe NPDR: any of $>$20 intraretinal haemorrhages in each of 4 quadrants, definite venous beading in 2 or more quadrants, or prominent IRMA in 1 or more quadrants, and no signs of proliferative retinopathy; \texttt{dme\_center\_involving\_od}: No macular oedema, or oedema sparing the central subfield; \texttt{dme\_center\_involving\_os}: Macular oedema involves the central subfield; \texttt{referable\_dr}: Refer to an ophthalmologist
}

\exbox{%
\textbf{Clinical activity score.} 55-year-old man, smoker; Graves' disease, radioiodine 8 months ago; attends thyroid eye clinic. RE more affected. RE: no spontaneous retro-orbital pain; no pain on eye movements; no eyelid redness; no conjunctival redness; no eyelid swelling; chemosis present; caruncle and plica not swollen; reduced up-gaze. Hertel exophthalmometry 22 mm RE, 19 mm LE.\\
\textbf{Questions and labels.} \emph{What is the 7-item Clinical Activity Score of the right eye (1 point each: spontaneous retrobulbar pain, pain on eye movement, eyelid redness, conjunctival redness, eyelid swelling, chemosis, caruncle/plica swelling)?} $\rightarrow$ CAS 1/7; \emph{Is the thyroid eye disease in the right eye active (7-item Clinical Activity Score of 3 or more)?} $\rightarrow$ Inactive disease (CAS 0 to 2)
}

\exbox{%
\textbf{Retinopathy of prematurity (ETROP).} Preterm infant born at 26 weeks' gestation, birth weight 1320 g, underwent ROP screening examination at 33 weeks postmenstrual age by dilated indirect ophthalmoscopy. In the right eye, retinal vessels end in zone I; there is a flat demarcation line, and the posterior pole vessels are of normal calibre and course. In the left eye, retinal vessels end in zone II; there is a ridge with extraretinal fibrovascular proliferation, and plus disease is present, with dilation and tortuosity of the posterior pole vessels.\\
\textbf{Questions and labels.} \emph{By the Early Treatment for ROP (ETROP) criteria, how should the right eye be managed now?} $\rightarrow$ Type 2 ROP: no treatment yet; close serial examination; \emph{By the Early Treatment for ROP (ETROP) criteria, how should the left eye be managed now?} $\rightarrow$ Type 1 ROP: treatment indicated now (ETROP); \emph{Does either eye meet the ETROP criteria for treatment now (type 1 ROP)?} $\rightarrow$ Treat now
}

\exbox{%
\textbf{Endophthalmitis (EVS).} An 85-year-old man presented 2 weeks after an intravitreal injection in the right eye. In the right eye he reported moderate pain and decreasing vision. Visual acuity in the right eye was hand motions. Examination showed conjunctival injection, anterior chamber cells and fibrin, and vitritis. There was no hypopyon. Iris details were visible through the cornea. There was no view of the fundus. B-scan showed vitreous opacities with an attached retina.\\
\textbf{Questions and labels.} \emph{Applying the Endophthalmitis Vitrectomy Study, what is the initial management of the right eye?} $\rightarrow$ Outside the EVS population (not acute endophthalmitis within 6 weeks of cataract or secondary IOL surgery); the EVS rule does not directly apply
}

\exbox{%
\textbf{Orbital infection (Chandler).} 59-year-old man with fever and a swollen, red right upper and lower eyelid. Right eyelid swelling is marked; lids cannot be opened unaided. Globe position is symmetric, with no proptosis. Right ocular motility is restricted and painful. Visual acuity is 6/6 in each eye, with no relative afferent pupillary defect. CT orbits with contrast shows a rim-enhancing fluid collection between the medial orbital wall and the periorbita, adjacent to an opacified ethmoid sinus. Left eye examination is normal.\\
\textbf{Questions and labels.} \emph{Classify this orbital infection on the Chandler classification.} $\rightarrow$ Group III: subperiosteal abscess
}

\exbox{%
\textbf{Visual impairment (WHO / ICD-11).} Dictated note. This 53-year-old man attended the vision assessment clinic. He wears his usual distance spectacles. Right eye: presenting distance acuity with current spectacles is counting fingers at 1 metre. Left eye: presenting distance acuity with current spectacles is logMAR 1.0; with pinhole, it is logMAR 0.6. Visual fields are full to confrontation in both eyes. Findings are grouped by eye.\\
\textbf{Questions and labels.} \emph{Classify the visual impairment by the WHO / ICD-11 categories, using presenting distance acuity in the better eye (and the central visual field).} $\rightarrow$ Moderate visual impairment: worse than 6/18, 6/60 or better
}

\exbox{%
\textbf{Anterior chamber cells (SUN).} Dictated note. 69-year-old woman with juvenile idiopathic arthritis-associated uveitis of the right eye, reviewed 3 weeks after the previous visit. RE: At the previous visit, 10 cells counted in a 1 mm by 1 mm slit beam. Today, 5 cells counted in a 1 mm by 1 mm slit beam; moderate flare, iris details clear. Vitreous clear at both visits. LE: Today, 2 cells counted in a 1 mm by 1 mm slit beam.\\
\textbf{Questions and labels.} \emph{Grade today's anterior chamber cells in the right eye on the SUN scale (cells counted in a 1 mm by 1 mm slit beam).} $\rightarrow$ Grade 0.5+: 1-5 cells; \emph{By the SUN activity definitions, how has the right eye's anterior chamber inflammation changed since the previous visit?} $\rightarrow$ Neither worsening nor improved by the SUN definitions
}

\end{document}